\documentclass[journal]{IEEEtran}

\usepackage{amsmath,bm, amssymb, amsfonts,color, caption, url, mathrsfs}
\usepackage{graphicx}
\usepackage[english]{babel}
\usepackage{multirow}

\ifCLASSINFOpdf
\else
\fi
\begin{document}
%
\title{SparseFusion: Dynamic Human Avatar Modeling from Sparse RGBD Images}
%
%
%

\author{Xinxin~Zuo~\IEEEmembership{Member,~IEEE}, Sen~Wang, Jiangbin~Zheng, Weiwei Yu, Minglun~Gong,~ Ruigang~Yang,~\IEEEmembership{Senior~Member,~IEEE} and~Li~Cheng,~\IEEEmembership{Senior~Member,~IEEE}, 
\IEEEcompsocitemizethanks{
\IEEEcompsocthanksitem Manuscript received ; revised ; accepted ... (L. Cheng and J. Zheng are co-corresponding authors for the paper.)
\IEEEcompsocthanksitem X. Zuo and S. Wang are with Northwestern Polytechnical University, Xi'an, 710072, China; University of Kentucky, Lexington, KY, 40508, USA; University of Alberta, Edmonton, AB, Canada.  (E-mail: xinxin.zuo@uky.edu, wangsen1312@gmail.com)
\IEEEcompsocthanksitem J. Zheng and W. Yu are with Northwestern Polytechnical University, Xi'an, 710072, China. (E-mail: {zhengjb, yuweiwei}@nwpu.edu.cn)
\IEEEcompsocthanksitem M. Gong is with University of Guelph, Guelph, ON, Canada. (E-mail: minglun@uoguelph.ca)
\IEEEcompsocthanksitem R. Yang is with University of Kentucky, Lexington, KY, 40508, USA. (E-mail: ryang@cs.uky.edu)
\IEEEcompsocthanksitem L. Cheng is with University of Alberta, Edmonton, AB, Canada. (E-mail: lcheng5@ualberta.ca)
}
}

\markboth{IEEE TRANSACTIONS ON MULTIMEDIA,~Vol.~XX, No.~XX, Month~Year}%
{Shell \MakeLowercase{\textit{et al.}}: Bare Demo of IEEEtran.cls for IEEE Journals}
%

\maketitle

\begin{abstract}
In this paper, we propose a novel approach to reconstruct 3D human body shapes based on a sparse set of RGBD frames using a single RGBD camera. We specifically focus on the realistic settings where human subjects move freely during the capture. The main challenge is how to robustly fuse these sparse frames into a canonical 3D model, under pose changes and surface occlusions. This is addressed by our new framework consisting of the following steps. First, based on a generative human template, for every two frames having sufficient overlap, an initial pairwise alignment is performed; It is followed by a global non-rigid registration procedure, in which partial results from RGBD frames are collected into a unified 3D shape, under the guidance of correspondences from the pairwise alignment; Finally, the texture map of the reconstructed human model is optimized to deliver a clear and spatially consistent texture. Empirical evaluations on synthetic and real datasets demonstrate both quantitatively and qualitatively the superior performance of our framework in reconstructing complete 3D human models with high fidelity. It is worth noting that our framework is flexible, with potential applications going beyond shape reconstruction. As an example, we showcase its use in reshaping and reposing to a new avatar. 
\end{abstract}


\IEEEpeerreviewmaketitle

%
\IEEEpeerreviewmaketitle

\section{Introduction}

\IEEEPARstart{3}D modeling or reconstruction of human bodies is an important topic that has wide range of applications in areas such as virtual reality, gaming, virtual try on, and teleconference. Many scanning systems under multi-view setup~\cite{Tong12Scanning, Alexiadis13Real, Dou13Scanning,dou2016fusion4d,Vlasic09Multiview} have been developed over the years, from which impressive results have been achieved. Such a system, on the other hand, is usually not portable and could be rather expensive. Rather than building on these sophisticated setups, in this paper we propose to reconstruct complete 3D human body shapes from a sparse set of frames taken by a single commodity-level RGBD camera. It is a challenging task especially in the presence of non-rigid articulated motions and surface occlusions.

The problem of recovering 3D models of deformable objects from a single depth camera has recently been studied. As an extension to the celebrated KinectFusion~\cite{Newcombe11KinectFusion} system, a dynamic fusion~\cite{Newcombe15DynamicFusion} approach has been developed which takes non-rigid motion into account by solving a non-rigid warp field for every frame.
However, they cannot handle fast motion, and the tracking error would accumulate as the sequence proceeds. To address these issues, several follow-up systems have been proposed to exploit either sparse feature correspondences~\cite{Innmann16voluedeform}, dense color information along the sequence~\cite{Guo17Real}, or the articulated motion constraints~\cite{Yu17bodyfusion,Yu18doublefusion} for more robust tracking, and enforcing loop closure~\cite{Dou15scanning, Wang18Dynamic} to recover a complete shape. 
The improved performance is achieved with a cost -- they rely on the existence of both a continuous image sequence and a reliable and continuous dense tracking over the entire sequence, which is computationally expensive and contains much redundant information. 
To account for this issue, we propose to instead consider only a sparse set of RGBD frames as input. The most related work is that of Li~\cite{Hao13selfportrait} and Shapiro ~\cite{shapiro2014rapid}, which takes several frames from a RGBD camera as input. However, in previous works, the user has to maintain a certain static pose while rotating in front of the camera, which is difficult to hold in practical settings. On the contrary, our proposed approach is capable of handling situations where human subjects are allowed to have significant pose changes.

To achieve this goal, we exploit the Skinned Multi-Person Linear model (SMPL)~\cite{Loper15SMPL} as a generative human template to register sparse frames of the human subject into a canonical model. First, the SMPL parameters are optimized to closely fit to the partial scans generated from the input depth images. Then, for every two partial scans that have sufficient overlap they are aligned by the correspondences conveyed and transferred via the SMPL template model. Starting from this pairwise alignment, a global non-rigid registration procedure is performed to get all those partial pieces deformed into canonical coordinate as guided by those correspondences acquired from the pairwise registration. After obtaining the 3D body shapes, a texture optimization approach is proposed to attach clear and consistent texture maps to the 3D model. During the texturing process, we take the non-rigid deformation into account, and deal with the possible misalignment by computing a warping field for each image successively. 



The proposed approach is examined on both synthetic and several real datasets captured with a single depth sensor. As demonstrated by the experiments, our approach is capable of generating complete and high quality human avatars from a very sparse set of RGBD frames.

The main contribution of this paper is that instead of taking a continuous depth sequence as input to fuse the sequence into a canonical model, we propose to use sparse RGBD frames to reconstruct a complete human avatar free from accumulation error. To be different from previous 3D self-portrait methods which usually assume static poses during the capture, we allow large pose variations by exploiting a statistical human template for the registration. 

As an interesting application, we can synthesize the reconstructed avatar by changing its shape and pose. A personalized SMPL model is built from the reconstructed human avatar. To achieve this goal, we propose a hierarchical representation of the reconstructed model with sparse control vertices mapped to the SMPL template, and the deformation of the reconstructed surface mesh is driven by those vertices. In this way, we could take advantage of the SMPL model in expressing human poses and shapes while still maintaining the surface details of the reconstructed model.

\section{Related Literature}
In this section, we review the related efforts on human body modeling. They could be roughly partitioned by the input modality and whether any human template is involved in the reconstruction.

\subsection{Human modeling with color images}
\label{Related:color}
The problem of 3D human body reconstruction has been studied for decades under the multi-view stereo setup~\cite{Wu11fusing, Aguiar08,zhu2016video} where multiple color images are taken as input. Typically, they exploit both the correspondence cues between images of neighboring views and the temporal consistency along the sequence to build up the involving surface. The involved multiple cameras are supposed to be synchronized and calibrated. Although very impressive and pleasing results have been achieved, this controlled setup is therefore mostly suitable in a laboratory setting. 

On the other hand, recent monocular human modeling approaches~\cite{NIPS2017Tung,varol2018bodynet,omran2018neural,alldieck2019learning,alldieck2018detailed,huang2018deep,natsume2019siclope,saito2019pifu,kolotouros2019convolutional} have shown compelling reconstruction results of human bodies from images in the wild. For example, Kanazawa et al.~\cite{kanazawa2018end} proposed an end-to-end framework to directly regress the parameters of a statistical body template from a single color image. 
A number of follow-up efforts proceed to incorporate additional information including body silhouettes, shading information ~\cite{natsume2019siclope,zhu2019detailed,alldieck2019learning}, or mutual constraints across multiple images~\cite{liang2019shape,huang2018deep} to train a neural network. Another branch of investigation is to employ volumetric representations~\cite{varol2018bodynet,zheng2019deephuman}, depth maps~\cite{tang2019neural} or UV maps~\cite{alldieck2019tex2shape} for the deep neural network. For instance, BodyNet~\cite{varol2018bodynet} learned to directly generate a voxel representation of the person using a deep neural network. However, due to the high memory requirements of voxel representations, fine-scale details are often missing in the output. 
Instead, PIFu~\cite{saito2019pifu} regressed an implicit surface representation that locally aligned pixels with the global context of the corresponding 3D object. Unlike voxel-based representations, this implicit per-pixel representation is more memory efficient.
Despite the widespread usage of learning based methods, the reconstructed human body usually lacks sufficient surface details. More importantly the inherent depth ambiguity of the color image stops the reconstructed human body from fitting closely to the real surface.



\subsection{Human modeling with depth images}
The advent of affordable consumer grade RGB-D cameras has brought about a profound advancement of human modeling approaches. There are some methods~\cite{Newcombe15DynamicFusion,Innmann16voluedeform,Guo17Real,Yu17bodyfusion,Yu18doublefusion} that use only a single depth sensor for the non-rigid objects reconstruction. As for the fusion based approaches, the surface is reconstructed in an incremental manner by tracking each frame along the RGBD sequence and updating the canonical model. First, as an extension to the KinectFusion system~\cite{Newcombe11KinectFusion}, a dynamic fusion approach~\cite{Newcombe15DynamicFusion} has been proposed to handle non-rigid motion by solving a non-rigid warp field for every frame. Later on, sparse feature information~\cite{Innmann16voluedeform} and dense color correspondences~\cite{Guo17Real} in the sequence were incorporated to improve the robustness of surface tracking. Besides, Yu et al.~\cite{Yu17bodyfusion} enforced the skeleton constraints in the typical fusion pipeline to get better performance on both surface fusion and skeleton tracking. Later on, a more robust fusion approach~\cite{Yu18doublefusion} was proposed by tracking both the inner and outer surface but they assume A-pose as the starting pose. Those methods allow the user to move more freely. However, as the sequence proceeds the almost inevitable drifting problem makes it difficult to recover a complete model without loop closure. 

To tackle the above mentioned problem and build up 3D self-portraits, there are efforts~\cite{Dou15scanning,Hao13selfportrait,Tong12Scanning,shapiro2014rapid,cui2012kinectavatar,mao2017easy,lin2016fast} that generate partial pieces in the first place and handle the error accumulation problem with a global registration. For instance, Shapiro et al.~\cite{shapiro2014rapid} aligned depth images from four static poses taken at 90 degree angles relative to each other with their proposed piecewise rigid registration method. Similarly, Li et al.~\cite{Hao13selfportrait} had eight partial scans as input and registered them globally with a non-rigid deformation approach. Mao et al.~\cite{mao2017easy} have taken 18 depth frames as input for the human modeling. However, they always assume static and same poses during capture. To make sure the pose is kept as same as possible during the capture, a turn-table was used in \cite{lin2016fast}. On the other hand, Dou et al.~\cite{Dou15scanning} allowed more free movement and proposed a non-rigid bundle adjustment method to align the partial pieces. Although impressive results were obtained, the bundle adjustment could be quite computationally expensive and time-consuming due to the large number of unknowns and search space.

Using a single depth sensor for human modeling is challenging as we need to handle the occlusion problem and the non-rigid motion. To meet this challenge, multiple depth sensors were exploited for dynamic surface modeling~\cite{Dou13Scanning,dou2016fusion4d}. For example, as the current state-of-the-art approach, Fusion4D~\cite{dou2016fusion4d} proposed a system for live multi-view performance capture, generating temporally coherent high-quality reconstructions in real-time. Although surfaces with great details have been reconstructed, the system is rather expensive and again takes extra effort to calibrate and synchronize the sensors. 

\subsection{Template-based human body modeling}

For the human body modeling, the idea of incorporating the human template has also attracted much attention. Early models were based on simple primitives~\cite{metaxas1993shape,gavrila19963}. The recent statistical human body models, such as SCAPE~\cite{anguelov2005scape} and SMPL model~\cite{Loper15SMPL}, were learned from thousands of scans of human bodies. The pose and shape deformations are encoded in the parametric model. Therefore, instead of recovering the 3D vertices on the surface, researchers~\cite{Bogo15Detailed, Zhang14Quality} set to obtain the pose and shape coefficients of the statistical model. For instance, a SCAPE based parametric human model was used in \cite{Bogo15Detailed} with a displacement map to represent the skin details. However, they did not take the surface deformation caused by cloth into account but assumed that the captured human subject is almost naked. In paper \cite{fechteler2019markerless,achenbach2017fast}, a kinematic skinning model was used for human pose and shape reconstruction from the 3D point cloud acquired by multi-view stereo methods. Alldieck et al.~\cite{alldieck2018detailed,alldieck2018video} took a monocular video sequence as input and exploited the SMPL model for coarse shape and pose estimation, together with the human silhouettes and image shading information for more detailed reconstruction. As we have reviewed in the learning based approaches in section \ref{Related:color}, the parametric human template also plays an important role in the recent learning based approaches, as only a small number of parameters are needed for regression. 

Instead of employing a general human template, there are endeavors~\cite{habermann2019livecap,xu2018monoperfcap,Zollhofer14real,Guo15Robust} that take pre-scanned human models as template for human performance capture. They are more related to surface tracking and the problem becomes easier to handle as the overall shape is already available. Furthermore, Yu et al. \cite{yu2019simulcap} also incorporated cloth simulation during the tracking procedure to model the deformation of inner body and outer cloth separately. 


In general, the template-based approaches are more reliable in handling occlusions, complex motion, and work well when the input is limited input such as a single or few images. In this paper, we utilize a probabilistic human template model to achieve more robust fusion under large pose changes, but still retain the surface details in the reconstructed model by using free-form deformation similar to the template-free approaches.

\section{Approach}
We are given sparse frames captured with a human subject under different poses with different body orientation. Therefore, for each frame we have a partial scan of the human body and our goal is to build up a complete model by fusing all those partial scans. In the following equations, $M_1 \sim M_N$ denotes the partial scans obtained from the depth images and $I_1 \sim I_N$ are the corresponding color images. In this paper, the SMPL model~\cite{Loper15SMPL} is used to register sparse frames into a canonical model.

The SMPL model is a skinned vertex-based model which parametrizes a triangulated mesh by pose and shape parameters. The shape parameters $\bm{\beta}$ are coefficients of a low-dimensional shape space, learned from a training set of thousands of registered 3D human body scans. The pose parameters $\bm{\theta}$ represent the joint angle in an axis-angle representation of the relative rotation between body parts. The posed body model $\mathscr{M}(\bm{\beta},\bm{\theta})$ is formulated as below given the shape and pose parameters,

\begin{equation}
\mathscr{M}(\bm{\beta},\bm{\theta}) = W(T_{P}(\bm{\beta}, \bm{\theta}), J(\bm{\beta}), \bm{\theta}, \Omega)
\end{equation}

\begin{equation}
T_{P}(\bm{\beta}, \bm{\theta}) = T + B_{S}(\bm{\beta}) + B_{P}(\bm{\theta})
\end{equation}
where $T$ is the base template mesh, $B_{S}(\bm{\beta})$ and $B_{P}(\bm{\theta})$ are vectors of vertices representing offsets from the base template as controlled by the shape and pose parameters respectively. Therefore, $T_p$ is the mesh of base template with the addition of both shape $B_{S}(\bm{\beta})$ and pose blend shapes $B_{P}(\bm{\theta})$. $J(\bm{\beta})$ is the joints position under the rest pose as controlled by the shape parameters. $W()$ is a blend skinning function which transforms the mesh from $T$ pose to the current pose $\bm{\theta}$ as controlled by the blending weights $\Omega$. More details about the SMPL model can be found in paper~\cite{Loper15SMPL}.
\vspace{5pt}

\begin{figure}[!]
	\centering
	\includegraphics[width=0.9999\linewidth]{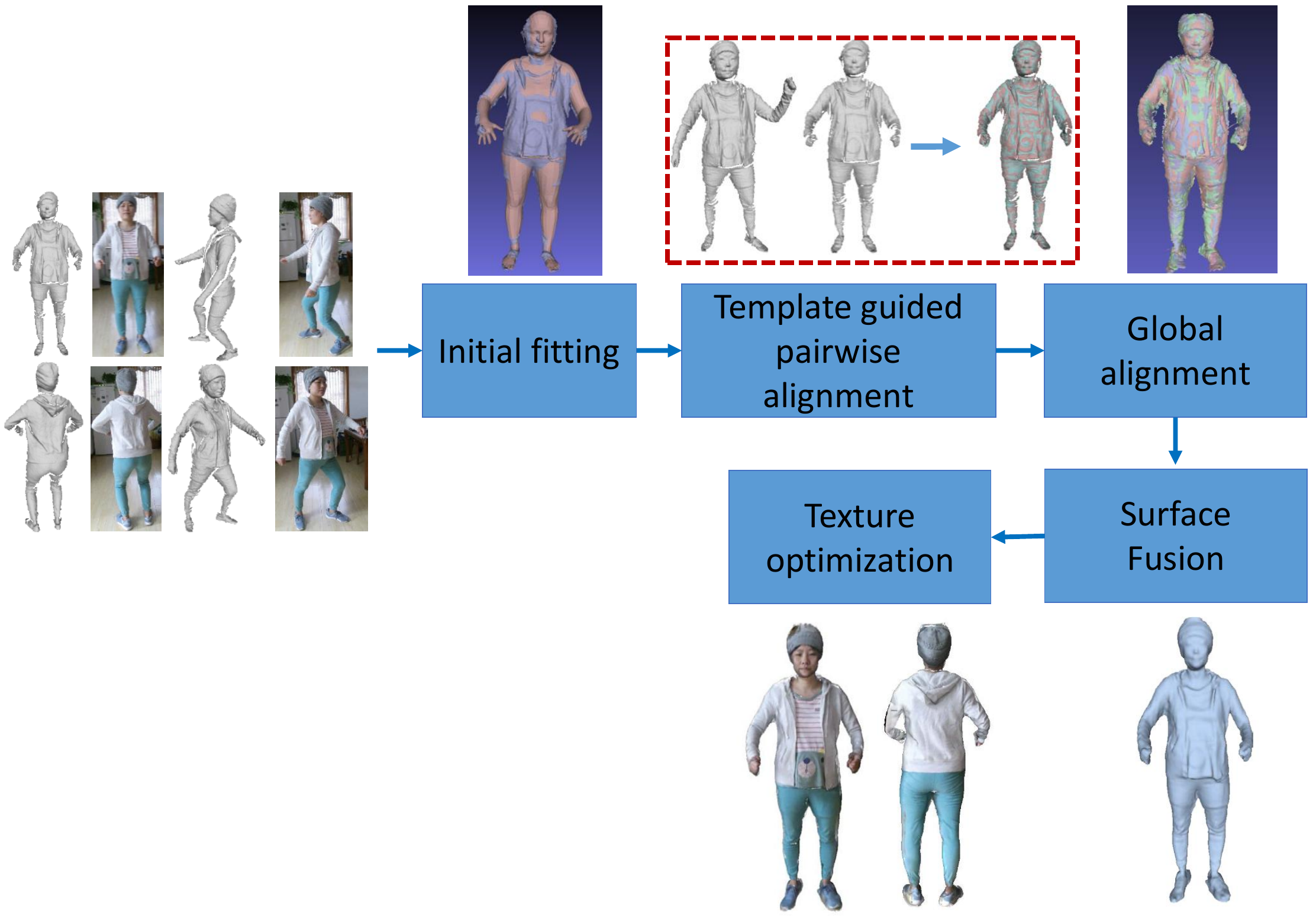}
	\caption{System Pipeline.}
	\label{Fig:pipeline}
\end{figure}

An overview of our method is shown in Figure~\ref{Fig:pipeline}. First, we optimize the SMPL model to let it fit to each of the partial scan. Afterwards we align every two partial pieces that have great overlap region by using the correspondences conveyed by the SMPL model. Finally, we register those pieces altogether with a global non-rigid registration approach. The model is further textured with our texture mapping procedure as described in Section~\ref{Sec:TexOpt}.

\subsection{Initial fitting} 
\label{Initial}

For every frame of the RGBD images, we solve the pose $\bm{\theta}$ and shape parameters $\bm{\beta}$ of the SMPL model so that the generated 3D human model fits as closely as possible to the captured RGBD image. For each frame $M_k$ and $I_k$, we achieve this by minimizing the following objective:
\begin{equation}
E(\bm{\beta}, \bm{\theta}) = E_{data}(\bm{\beta}, \bm{\theta}) + \alpha_{r} E_r(\bm{\theta})
\end{equation}

The data term $E_{data}$ is defined as: 
\begin{equation}
E_{data}(\bm{\beta}, \bm{\theta})  =  E_{surface}(\bm{\beta}, \bm{\theta})  + \alpha_{j} E_{joints}(\bm{\beta}, \bm{\theta}) 
\end{equation}

First, we have the surface fitting term $E_{surface}$ so that for each vertex $M_{k}^{i}$ in the surface $M_k$, we minimize its distance to the closest vertex on the generated SMPL model $\mathscr{M}(\bm{\beta},\bm{\theta})$:
\begin{equation}
E_{surface}(\bm{\beta}, \bm{\theta})  =  \sum_{i \in |\mathscr{M}_k|} \min_{v \in \mathscr{M}(\bm{\beta},\bm{\theta})} ||\mathscr{M}_k^i -v||_2^2
\end{equation}

The joints fitting term $E_{joints}(\bm{\beta}, \bm{\theta})$ is formulated to match the model joints to the joints of the partial scans (denoted as $\hat{J}_{est,i}$). $f()$ is the function that transforms the joint from its rest pose to current positions as controlled by the pose parameters using the chain rule defined by the human skeleton. We compute the 2D joint locations in the color image using OpenPose~\cite{Cao18Openpose}, after which the 3D human joints are estimated by back-projecting the 2D joints into 3D space with the depth information. $\rho()$ is a robust Geman-McClure penalty function~\cite{Geman87penalty}. This term is important to address large pose changes.

\begin{equation}
E_{joints}(\bm{\beta}, \bm{\theta}) = \sum_{i\in|J|}\omega_i \rho (f(J(\bm{\beta})_i, \bm{\theta})- \hat{J}_{est,i})
\end{equation}

The other term $E_r(\bm{\theta})$ is a pose regularization term formulated as below which penalizes unusual poses. It is defined as a Gaussian mixture model trained from the CMU dataset~\cite{CMUmocap19} where $N(\bm{\theta};\mu_{\bm{\theta},i},\Sigma_{\bm{\theta},i})$ is a Gaussian distribution with its mean and variance denoted as $\mu_{\bm{\theta},i}$ and $\Sigma_{\bm{\theta},i}$ respectively.

\begin{equation}
E_r(\bm{\theta}) =  -\log \sum_{i}(c_i N(\bm{\theta};\mu_{\bm{\theta},i},\Sigma_{\bm{\theta},i}))
\end{equation}

We get the shape and pose parameters for each piece by minimizing the above objective function so that the optimized SMPL model will fit to the partial scans.

Furthermore, for every partial scan they should have consistent body shapes as for the same human subject. Therefore, we propose a bundle adjustment approach to refine the shape and pose parameters by minimizing the total misalignment error of all those partial pieces to the SMPL model with respect to a consistent body shape and their poses respectively. Mathematically the objective function is formulated as below,


\begin{equation}
E(\bm{\Omega}, \bm{\beta}) =  \sum_{k=1}^{N}E_{surface}(\bm{\beta}, \bm{\theta}_k)
\end{equation}
\begin{equation}
\bm{\Omega}=  \{ \bm{\theta}_1, \bm{\theta}_2, \cdots \bm{\theta}_N\}
\end{equation}

\begin{figure}[!]
	\centering
	\includegraphics[width=0.9999\linewidth]{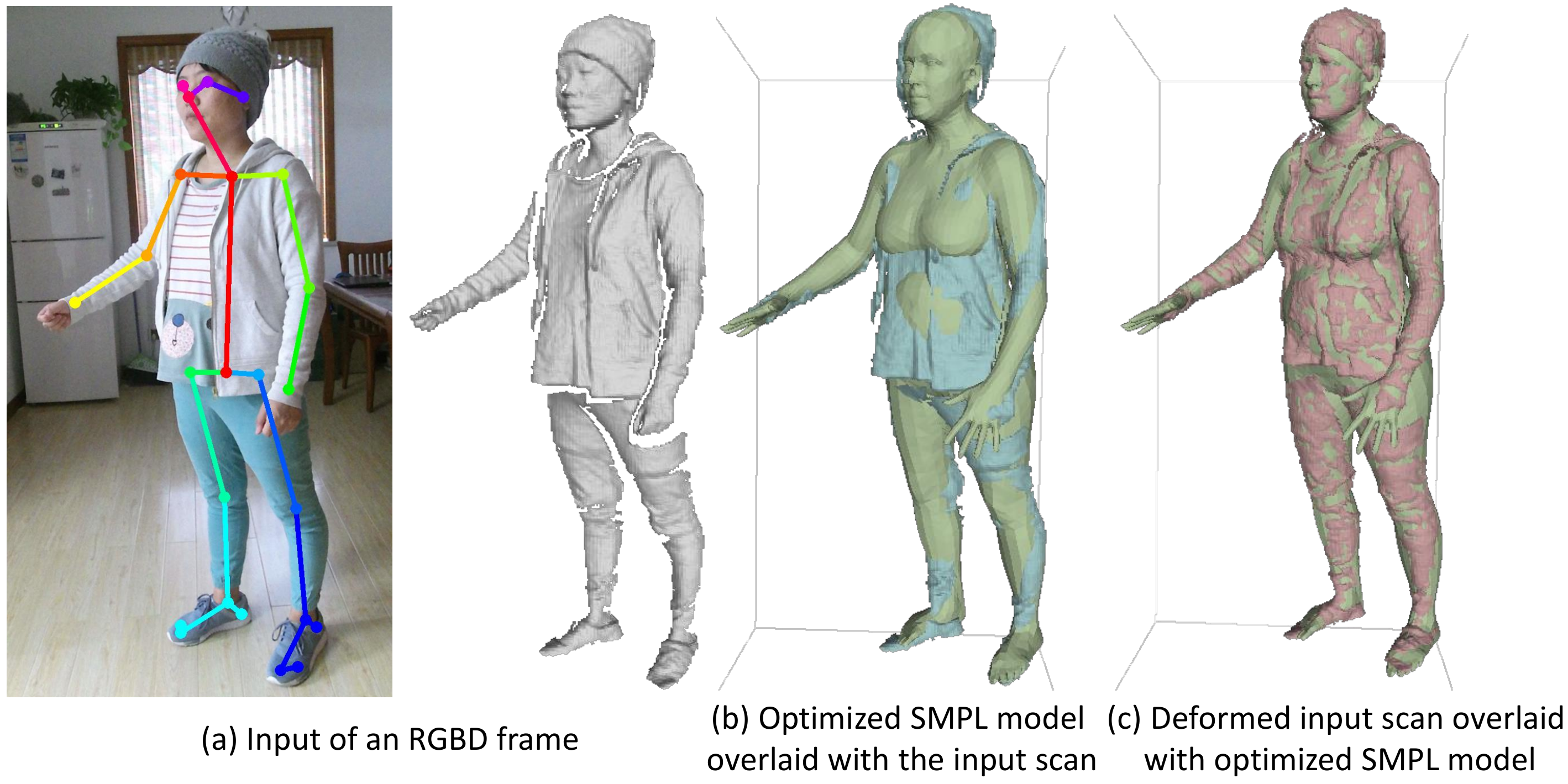}
	\caption{Initial Fitting results. (a) is the input RGBD frame and we show the detected joints on the color image. (b) shows the optimized SMPL aligned with the input scan. (c) shows the deformed input scan that fits even better to the SMPL model.}
	\label{Fig:initial}
\end{figure}

We initialize the pose parameters with those computed separately from each piece. The shape parameters are initialized by the one computed from a frontal piece.
We show the fitting results in Figure~\ref{Fig:initial} showing the optimized SMPL that fits to the input partial scans.

\subsection{Template guided pairwise alignment}
After we get the optimized SMPL model that fits to the input RGBD images, we take it as guidance for initial alignment of those partial scans. Before that, since we cannot find any SMPL model that will fit perfectly to the input mesh because of the casual clothes, we further deform the input mesh onto the optimized SMPL model to get better alignment, as shown in Figure~\ref{Fig:initial}(c). After that, we can establish correspondences from every input scan to the optimized SMPL model via nearest search. And then the correspondences between every two input scans are established via the SMPL model.


Similar to the registration approach proposed in ~\cite{li2008global}, we register partial scans by exploiting the Embedded Deformation Model (EDM) ~\cite{Sumner07EDM} to parametrize the mesh. To be different from the previous registration method which  requires the partial scans to be close to each other so as to have a proper initialization, we get the correspondences between the partial scans via the SMPL model. We describe the proposed method in detail below.

For the deformation model, a set of graph nodes ($g_1, g_2, ...,g_l$) are uniformly sampled throughout the mesh, and for each node $g_i$, it has an affine transformation specified by a $3*3$ matrix $\bm{A}_i$ and a $3*1$ translation vector $\bm{t}_i$. For each vertex $v$ on the mesh it is controlled and deformed by its $K$ nearest graph nodes with a set of weights:

\begin{equation}
\Phi(v) = \sum_{i=1} ^{K} w_i(v) [\bm{A}_i (v -g_i) +g_i +\bm{t}_i]
\end{equation}

We compute the deformation from $M_i$ to $M_j$ by building a graph for the mesh $M_i$ and estimate the deformation parameters $\bm{A}_1\sim \bm{A}_l$ (denoted as $\mathcal{A}$) and $\bm{t}_1\sim \bm{t}_l$
(denoted as $\mathcal{T}$) by minimizing the following objective function:
\begin{equation}
E(\mathcal{A},\mathcal{T}) = \alpha_{reg} E_{reg}(\mathcal{A}) + \alpha_{s} E_s(\mathcal{A},\mathcal{T}) + E_{cor}(\mathcal{A},\mathcal{T})
\end{equation}

The term $E_{reg}$ serves as the as-rigid-as-possible term preventing arbitrary surface deformation.
\begin{equation}
E_{reg}(\mathcal{A}) = \sum_{i=1} ^{l} || \bm{A}_{i} \bm{A}_{i}^{T} -I ||_{2}^{2}.
\label{Eq:Er}
\end{equation}

The smoothness term $E_s$ ensures smooth deformation of neighboring graph nodes.

\begin{equation}
E_s(\mathcal{A}, \mathcal{T}) = \sum_{(i,j)\in \mu}||\bm{A}_i(g_j-g_i) + g_i + \bm{t}_i -( g_j+ \bm{t}_j)||_{2}^{2}.
\label{Eq:Es}
\end{equation}

The term $E_{cor}$ is our data term which penalizes the distances between correspondences on these two pieces, which are extracted through the above optimized SMPL model $S_i$ for $M_i$ and $S_j$ for $M_j$. Specifically, for a vertex $v_p$ on piece $M_i$, we find its nearest vertex on $S_i$ within a certain threshold, which is denoted as $v_s$. And we extract the vertex from $S_j$ which has the same vertex index as $v_s$. Then we find the nearest vertex for $v_s$ with respect to the mesh $M_j$, which is denoted as $v_q$. The distance between $v_p$ and $v_q$ is minimized.
\begin{equation}
E_{cor}(\mathcal{A}, \mathcal{T}) = \sum_{(v_p,v_q)\in \mathcal{C}_{ij}}|| \Phi(v_p) - v_q||_{2}^{2}.
\label{Eq:cor}
\end{equation}

\begin{figure}[!]
	\centering
	\includegraphics[width=0.95\linewidth]{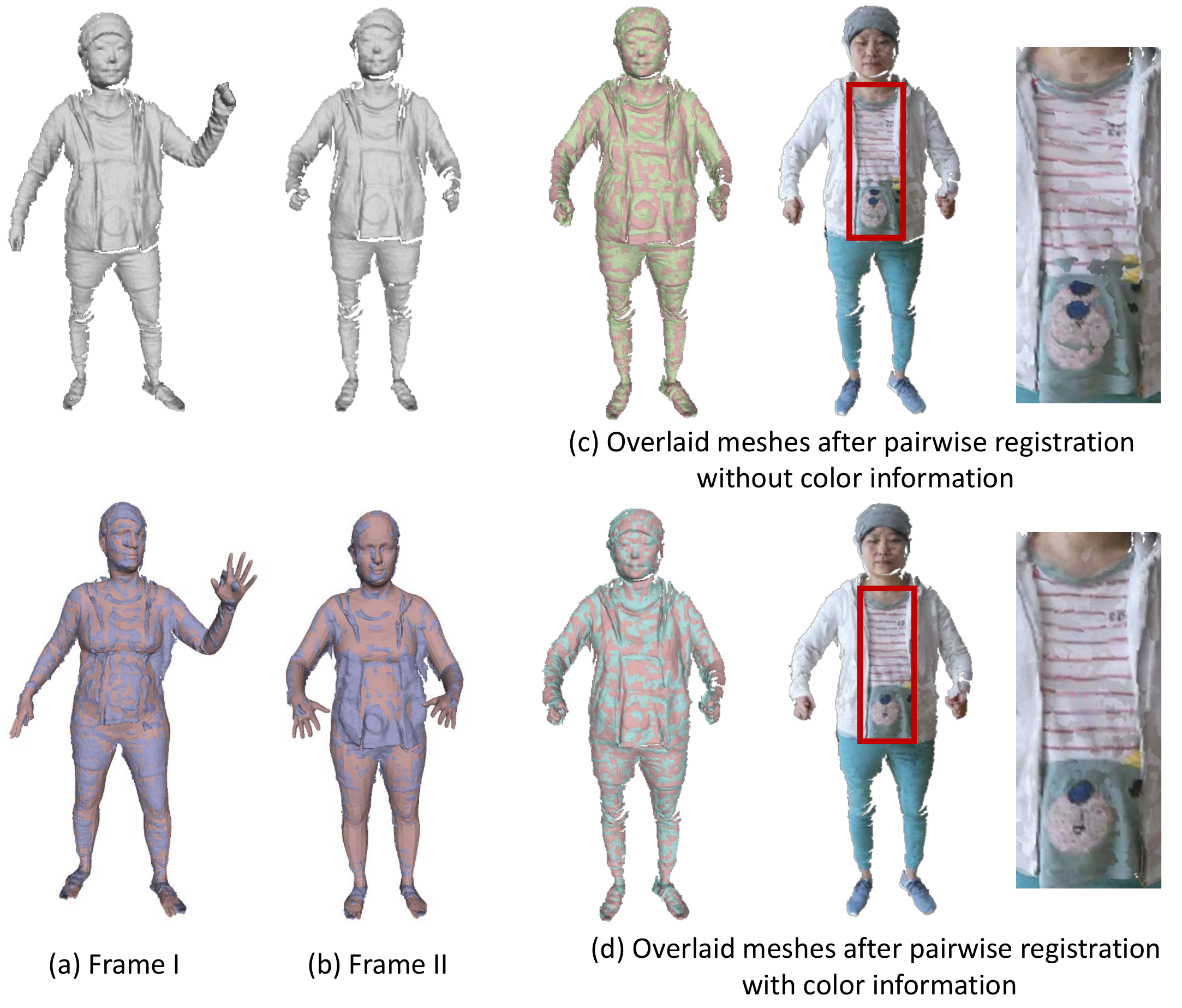}
	\caption{Pairwise registration results. (a) and (b) are two sampled pieces. We also demonstrate the overlay of optimized SMPL model and the input scans below. The mesh of (a) is deformed onto the mesh of (b). (c) shows our registration result of mesh (a) and mesh (b) but without color information. (c) shows our registration result of mesh (a) and mesh (b) with color information. We also display the overlaid meshes with color attached to demonstrate the effectiveness of the color information for the registration.}
	\label{Fig:pairwise}
\end{figure}

To get better alignment, we use the color information to refine the initial registration. In details, first every partial scan is textured with its corresponding color image. Suppose we have got the deformed mesh of $M_i$ which is aligned to $M_j$ after the above registration, and we denote it as $D_{i}^{j}$. Now, we render a color image $I_i$ with the deformed mesh $D_{i}^{j}$ onto the same space with respect to the color image $I_j$. We compute a flow field from $I_i$ to $I_j$ and map the flow correspondences to the meshes. Finally, the deformation from $M_i$ to $M_j$ is further optimized using the EDM by enforcing the color correspondences. We show a pairwise registration result in Figure~\ref{Fig:pairwise}. As shown in Figure~\ref{Fig:pairwise}, we are able to align pieces that have large pose variation. As we can see in Figure~\ref{Fig:pairwise}(c), it seems that we can already get good overlaid meshes without color information. However, the misalignment still exists which can be seen clearly when we attach color onto the meshes. Therefore, we enforce the color correspondences to resolve this issue where Figure~\ref{Fig:pairwise}(d) shows the overlaid meshes.

\begin{figure*}[!ht]
	\centering
	\includegraphics[width=0.9999\linewidth]{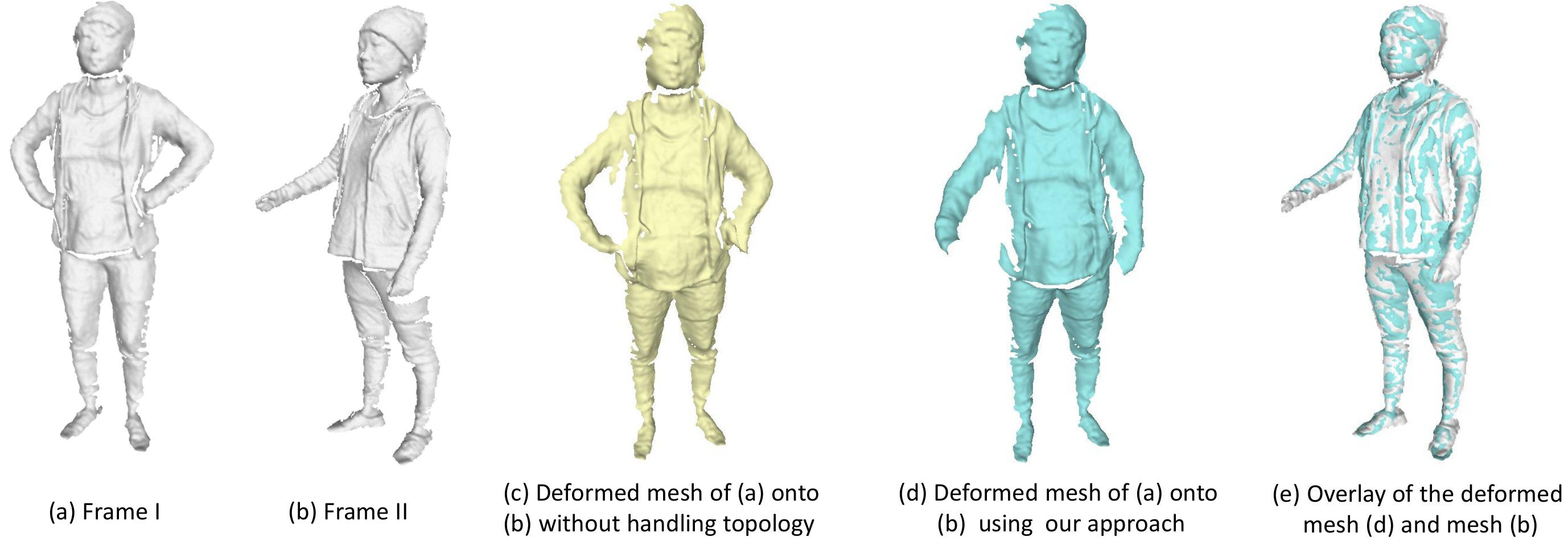}
	\caption{Pairwise registration results with topology changes. (a) and (b) are two sampled pieces. We try to deform mesh (a) onto mesh (b) where the topology has changed. (c) shows the deformed mesh of (a) without taking the topology change into account. (d) shows the deformed mesh using our approach. (e) is the overlay of deformed mesh and the target mesh. }
	\label{Fig:topology}
\end{figure*}

\textbf{Topology Change}
Another important property of our method on pairwise registration is that we are able to deal with the topology changes quite conveniently by exploiting the information provided by the human template. That is, we can extract body part information from the optimized template model and assign a body part for each vertex of the input mesh. First, we delete the faces for which their corresponding vertices do not belong to the same body part nor do they belong to the body parts that have parent or child relationship. Next, while building up the embedded graph, we only connect graph nodes that belong to the same body part or neighboring parts. In the meanwhile, we set further constraints that the vertex is controlled by the graph nodes belonging to either the same body part or neighboring parts defined by its parents or child nodes. We show example of pairwise registration of two partial pieces that have topology changes in Figure~\ref{Fig:topology}. As we want to deform mesh of Figure~\ref{Fig:topology}(a) to the mesh of Figure~\ref{Fig:topology}(b) where the topology has changed, the deformation cannot get implemented correctly without explicitly handling the topology change(Figure~\ref{Fig:topology}(c)). However, the problem can be resolved with our method by taking advantage of the semantic information contained in the template. The deformed mesh with our approach is shown in Figure~\ref{Fig:topology}(d), which aligns well with the target mesh as shown in Figure~\ref{Fig:topology}(e).

\begin{figure*}[!t]
	\centering
	\includegraphics[width=0.9999\linewidth]{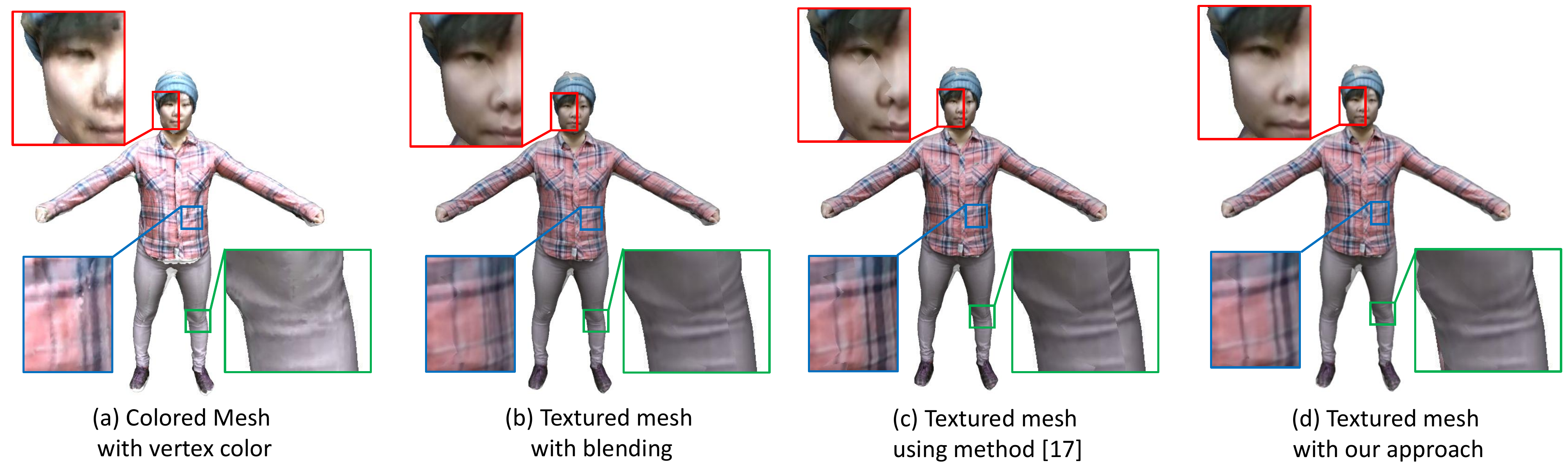}
	\caption{Texture mapping results.}
	\label{Fig:texOpt}
\end{figure*}

\subsection{Global alignment}
After the initial alignment, we are able to establish correspondences between those partial pieces, with which we can align them globally into a canonical model. Similar to the registration of two partial pieces, we exploit the Embedded Deformation Model here to extrapolate the deformation field. It means for every partial piece($M_1\sim M_N$) we have a deformation graph embedded with it and our goal will be to solve those graph parameters($\mathbb{A} = \mathcal{A}_1 \sim \mathcal{A}_N$, $\mathbb{T} = \mathcal{T}_1 \sim \mathcal{T}_N$) altogether. The objective function is formulated as,
\begin{equation}
\begin{split}
E(\mathbb{A},\mathbb{T}) &= \sum_{i=1}^{N} [\alpha_{reg} E_{r}(\mathcal{A}_i, \mathcal{T}_i) + \alpha_{s} E_{s}(\mathcal{A}_i, \mathcal{T}_i)] \\
&+\alpha_{corr} E_{corr}(\mathbb{A},\mathbb{T})
\end{split}
\label{Eq:sum_nonrigid}
\end{equation}


The first two terms are the as-rigid-as-possible and smoothness term respectively as defined in Equation~\ref{Eq:Er} and~\ref{Eq:Es}. We have the third term $E_{corr}$ defined as below as the data term enforcing the correspondences between partial scans achieved from the above pairwise initial alignment.
\begin{equation}
E_{corr}(\mathbb{A},\mathbb{T}) = 
\sum_{(M_s,M_r)\in \mathcal{U}}\sum_{(p_i,q_i)\in C_{sr}} ||\phi(M_s^{p_i},\mathcal{A}_s, \mathcal{T}_s) - M_{r}^{q_i}||_{2}^{2}
\label{Eq:corr}
\end{equation}

where $M_s$ and $M_r$ are any two pieces that have sufficient overlaps, and $C_{sr}$ is the correspondence set we have got after the pairwise alignment. The deformed mesh of $M_s$ is supposed to fit onto the target mesh $M_r$ as controlled by the correspondences. Besides, vertices of the reference frame is enforced as fixed constraints.

Finally, with all those input partial pieces deformed to a canonical space, we apply Poisson surface reconstruction and get the final fused human model.

\subsection{Texture optimization}\label{Sec:TexOpt}
In some applications such as free-viewpoint video generation and teleconference, a 3D geometric human body is not enough and we want the model to be textured. Previous human model scanning systems that use a single RGBD camera usually output models with per-vertex color since it is rather difficult to maintain and update the texture atlas during the fusion process. However, the per-vertex color could be very blurry (as shown in Figure ~\ref{Fig:texOpt}(a)) when the resolution of the mesh is not high enough. Therefore, instead of computing per-vertex color we attach texture maps onto the model.
The input is the reconstructed human model together with those partial pieces aligned to the canonical model as well as their corresponding color images. Our goal is to generate a consistent and clear texture map for the 3D human model given the input.

There are some texture mapping methods that project the meshes onto multiple image planes, and then adopt weighted average blending strategy to synthesize model textures. However, the generated texture is still blurry in our case (as shown in Figure ~\ref{Fig:texOpt} (b)) as the misalignment between those partial pieces still exists, which means the textures from different images are not perfectly matched. Previous approaches ~\cite{gal10seamless} tackled the misalignment problem by selecting the textures from multiple views while minimizing the seams. But it also failed in our case (as shown in Figure~\ref{Fig:texOpt} (c)) as we only have sparse input frames. Therefore, instead of directly synthesizing from multiple images, we try to eliminate possible misalignment and optimize a warping field for every image consecutively before attaching these to the mesh model. We describe our texture optimization approach below.

\begin{figure*}[!h]
	\centering
	\includegraphics[width=0.9999\linewidth]{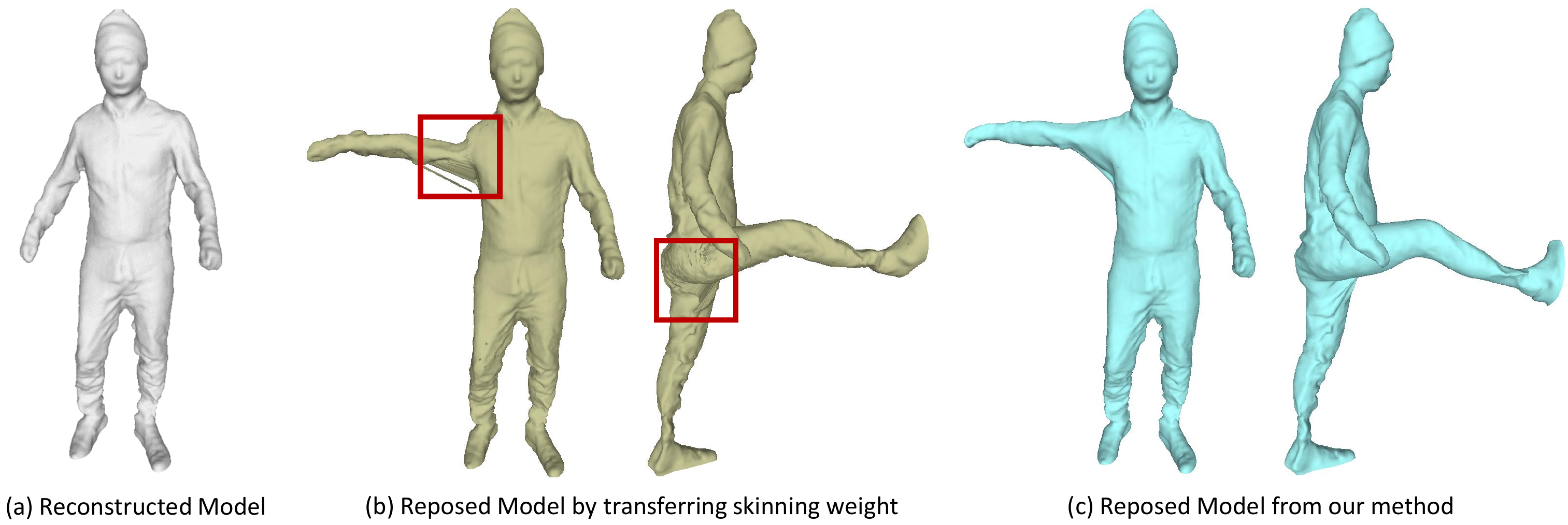}
	\caption{Comparison of reposing of a human avatar.}
	\label{Fig:reposecmp}
\end{figure*}

Starting from the reference frame, we attach the corresponding image onto the reconstructed mesh model by projecting the mesh onto the image plane and compute the texture coordinates for every face that is visible in the reference frame. For the next neighboring frame $k$, we deform the reconstructed human model onto mesh $M_k$ using the correspondences acquired from the above global registration. Then, we render a color image $I_{model}$ with respect to the view direction of frame $k$ from the current textured human mesh model. On the other hand, we have the captured color image $I_k$ for the frame $k$. The possible misalignment between $I_{model}$ and $I_k$ will cause visual seams if we attach the image $I_k$ directly onto the current human mesh. To address this problem, instead of adjusting the texture coordinates for each face in the 3D mesh which is difficult to optimize, we try to find a warping field $W_k$ for $I_k$ in the image plane so that the warped image will be well aligned with $I_{model}$. In details, first we detect the overlap regions of the texture map between $I_{model}$ and $I_k$, which we denote as $\Omega_o$. A flow field $\hat{W_k}$ is computed from $I_k$ to $I_{model}$ for the overlap part.
Next, we propagate the flow field to the non-overlap part $\Omega_N$ by minimizing the following objective function, from which the overall warping field $W_k$ is estimated,

\begin{equation}
\begin{split}
E(W_k) & = \sum_{p \in \Omega_o}||W_k(p) - \hat{W_k}(p)||^2 \\
&+ \lambda_s \sum_{(p,q )\in N}||W_k(p)-W_k(q)||^2 + \lambda_b \sum_{p \in \Omega_N}||W_k(p)||^2
\end{split}
\label{Eq:warpTex}
\end{equation}

where the first term is to keep the warping field close to the estimated flow filed in the overlap region and the second term is enforced to keep the warping field as smooth as possible so that we can propagate the flow to the non-overlap region. Finally, we introduce the last term as a boundary term to set constraints for pixels that are not connected to the overlap regions.

Afterwards, we select the optimal texture image for each face of the human model to generate the final texture maps. In Figure~\ref{Fig:texOpt}, we show the texture mapping results w/o our texture optimization procedure.

\section{Implementation details}

To capture the real dataset, we have used the Kinect V2 and the human subject is asked to rotate in front of the camera. But we do not assume any specific pose or slow motion during capture. We have captured twelve frames for each human subject. But we only used two or four frames in some case as demonstrated in \ref{Sec:Exp}. The captured depth maps are quite noisy, so they are smoothed in the first place as a preprocessing step before fusion.

The parameters $\alpha_{r}$, $\alpha_{j}$ in the initial fitting objective function are set to be 7.5, 2.0 respectively. For the deformation model, $\alpha_{reg}$ is set to be 0.2, $\alpha_{s}$ is 0.5 and $\alpha_{corr}$ is 1.0. For each input scan, we evenly sample 500 nodes over the mesh to build up the deformation graph. During the warping field computation in texture optimization process, $\lambda_{s}$ is set to be 0.8 and $\lambda_{b}$ is 1.0. Those parameters are manually tuned and kept fixed in all the experiments shown in the paper.

We implement most parts of our framework in Matlab. We run the algorithm on a desktop with 8-core 3.2GHz Intel CPU and 32 GB memory. It takes approximately 490s for the overall framework. In details, for the initial fitting, it takes about 14s for every piece and 116s for pairwise registration, and finally 107s for the global alignment. The texture mapping procedure takes about 104s.

\section{Applications}
\begin{figure}[b]
	\centering
	\includegraphics[width=0.9999\linewidth]{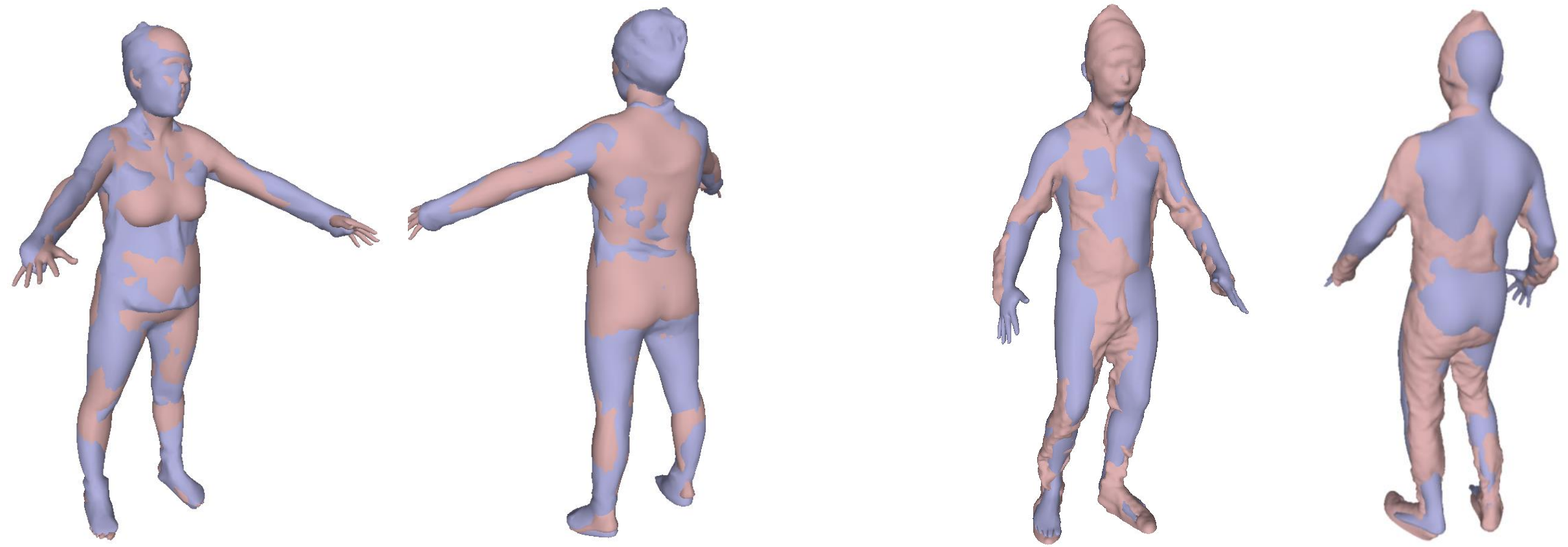}
	\caption{Illustration of our personalized avatar generation. We optimize the SMPL model to has a close fit to the reconstructed model before building up our personalized SMPL model.}
	\label{Fig:per-smpl}
\end{figure}
In this section we present an useful application to generate human models under various shapes and poses by building up a personalized SMPL model from the reconstructed human avatar estimated with our proposed sparse fusion approach.

\begin{figure*}[!ht]
	\centering
	\includegraphics[width=0.9999\linewidth]{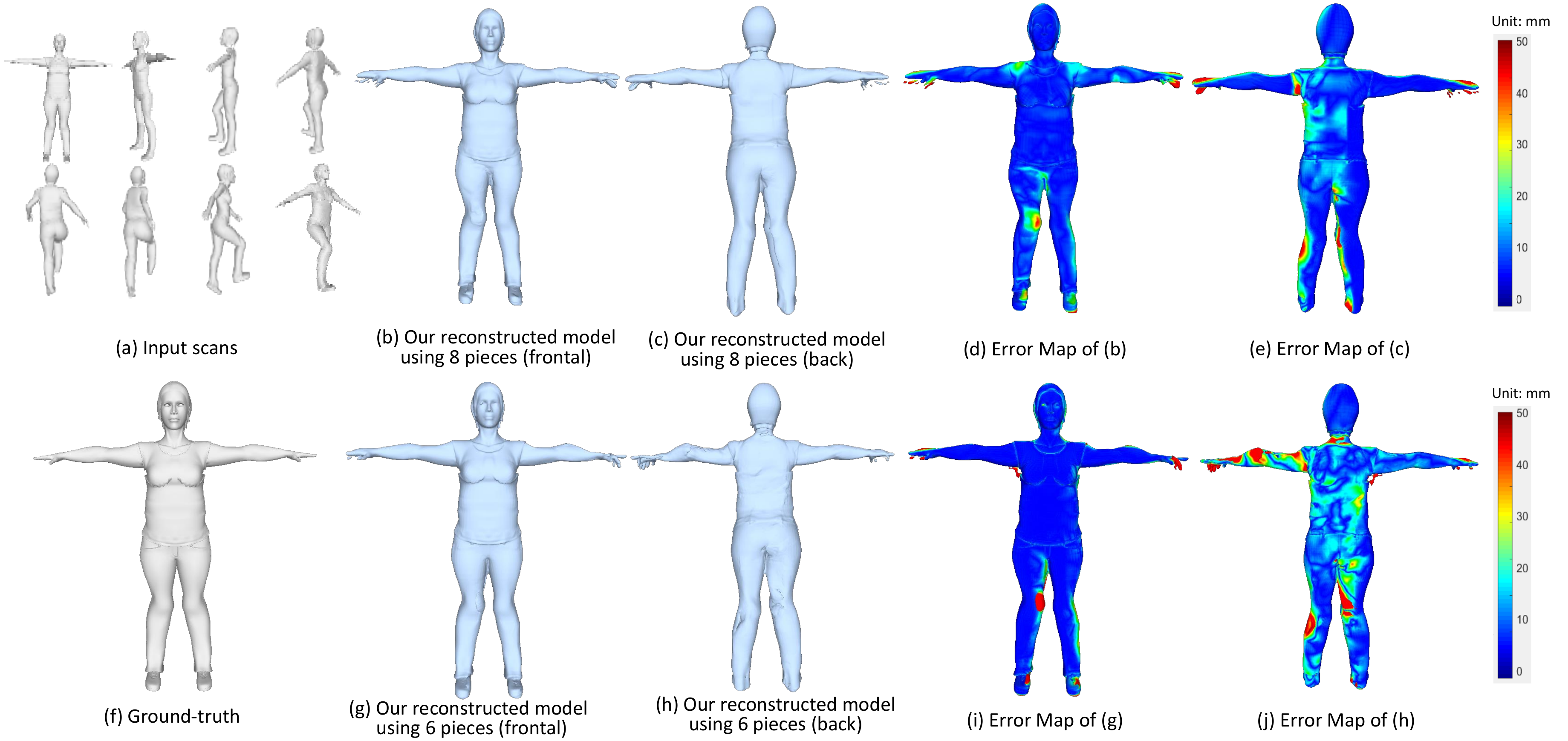}
	\caption{Results on a synthetic dataset.}
	\label{Fig:synthetic}
\end{figure*}

Previous approaches drive the human avatar via manual or auto rigging and setting up the skinning weights. However, it is not a trivial task to set proper skinning weights which will produce unrealistic deformations at the joints as shown in Figure~\ref{Fig:reposecmp}. In addition to reposing the reconstructed model, we want to be able to adjust the shapes and synthesize avatar to be fatter or thinner. This is not easy to achieve via the simple skinning weights transfer. Therefore, instead of transferring the skinning weights from a general template, in this paper we embed the SMPL model into the reconstructed human avatar and propose a hierarchical representation for deformation. That is, we want to take advantage of the SMPL model for body reshaping and reposing and also to preserve the surface details beyond the SMPL model.

Starting from the SMPL model we have got from the initial fitting procedure by fitting to those partial pieces as described in Section~\ref{Initial}, we further optimize it to have a closer fit to the complete 3D model after the fusion. This is achieved in a similar fashion to the initial model fitting. The only difference is we do not need to enforce any prior in this case as we already have a good initial model. Besides, the complete human model obtained through our SparseFusion method provides us with sufficient constraints for the estimation of the SMPL parameters. Therefore, we just need to penalize the distance between the SMPL model and the reconstructed human model by solving the objective function as defined in Equation~\ref{Eq:Es}. We show the optimized SMPL overlaid with the reconstructed model in Figure~\ref{Fig:per-smpl}.

In the next step, for each vertex in the SMPL model we could find its correspondence in the reconstructed model via nearest search. We construct a displacement map $S_d$ from the SMPL model to these correspondences on the reconstructed mesh. The SMPL model could be reposed or reshaped by setting up the pose or shape parameters. We apply the displacement map $S_d$ to the reposed mesh, which is denoted as $P(\bm{\beta},\bm{\theta})$. 

\begin{equation}
T_{P}^{d}(\bm{\beta}, \bm{\theta}) = T + B_{S}(\bm{\beta}) + B_{P}(\bm{\theta}) + S_d
\end{equation}

\begin{equation}
P(\bm{\beta},\bm{\theta}) = W(T_{P}^{d}(\bm{\beta}, \bm{\theta}), J(\bm{\beta}), \bm{\theta}, \Omega)
\end{equation}

However, the repose SMPL mesh still lacks surface details. We take it as intermediate mesh and the vertices on the mesh as control points to deform the reconstructed avatar under the as-rigid-as-possible deformation. The animation results are shown in Figure~\ref{Fig:reposecmp}(c).

\begin{figure}[t]
	\centering
	\includegraphics[width=0.99999\linewidth]{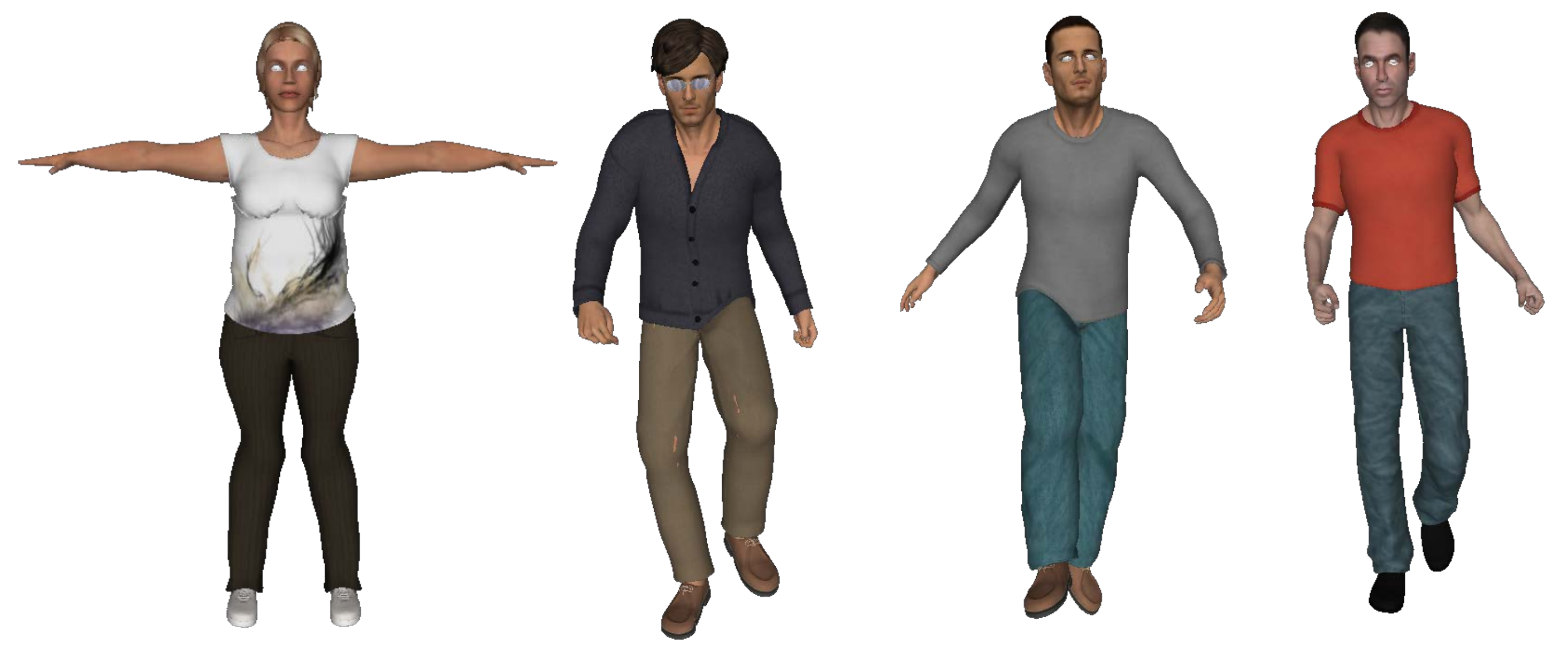}
	\caption{Models of synthetic datasets.}
	\label{Fig:synModel}
\end{figure}


\section{Experiments}
\label{Sec:Exp}
\begin{figure*}[!h]
	\centering
	\includegraphics[width=1.00\linewidth]{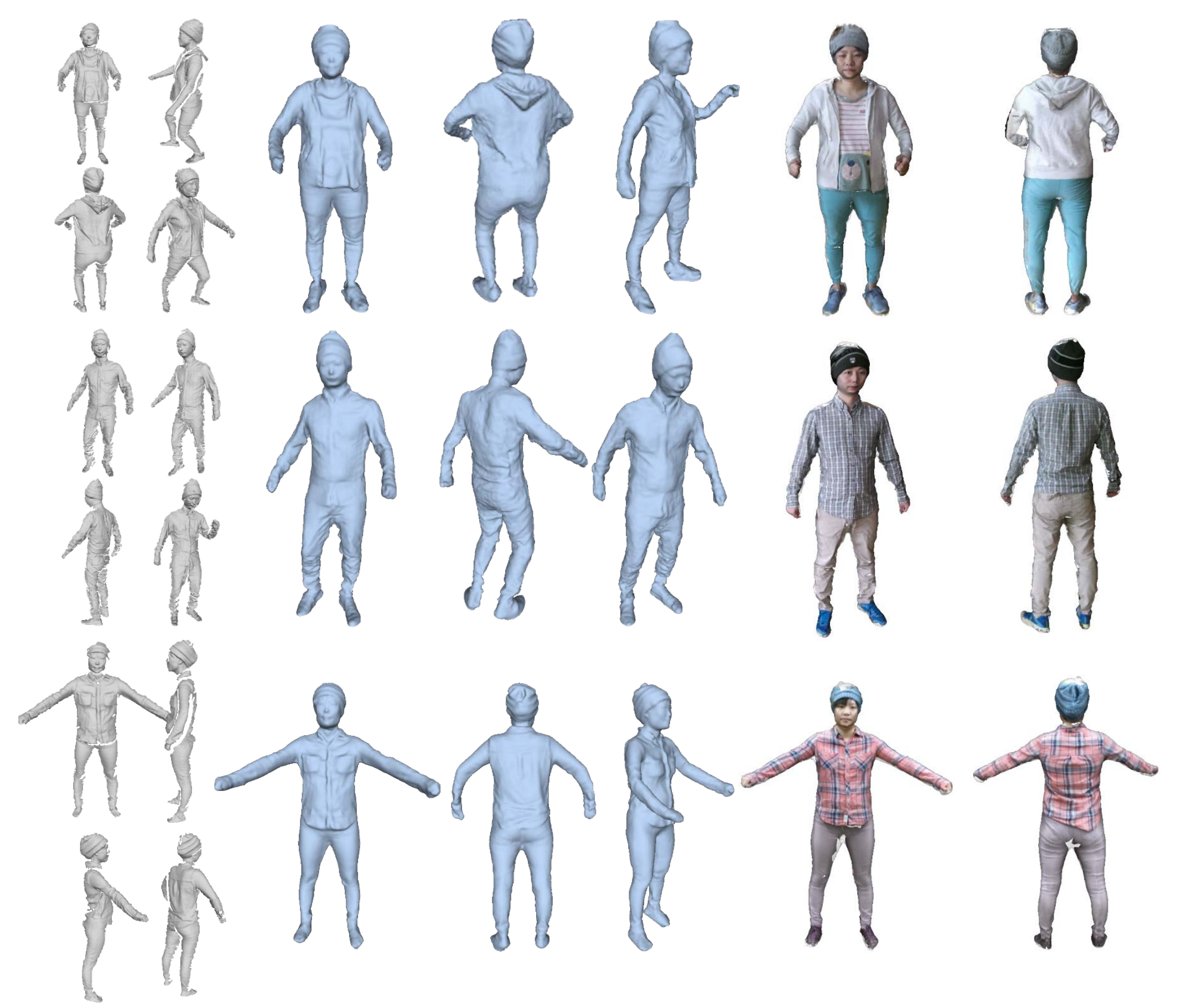}
	\caption{Results on real datasets. The left four columns are sampled input scans; The three middle columns are the fused model and models deformed to some input scans. We display the textured models in the two rightmost columns. The number of the vertices for the three reconstructed models from top to down is 60385, 54281 and 57826 respectively.}
	\label{Fig:real}
\end{figure*}

We demonstrate the effectiveness of our approach in the experimental part with both quantitative and qualitative results.

\subsection{Quantitative evaluation on synthetic datasets}
We tested our system on synthetic datasets that we have created using Poser~\cite{Poser19}. We have selected four human subjects (as shown in Figure~\ref{Fig:synModel}) and for each human subject we generate eight models under different poses. We synthesize one depth map and one color image for each model with a virtual camera rotating around the subject, which means we have got eight depth maps and color images as input with each frame corresponds to a model in a specific pose. We demonstrate an example in Figure~\ref{Fig:synthetic}(a). Our reconstruction system results in a shape (as shown in Figure~\ref{Fig:synthetic}(b)(c)) with respect to the first selected frame which is taken as the canonical frame. We plot the error map to show the geometric error of our reconstructed model with respect to the ground-truth model. The error for each vertex is computed via a nearest search to the ground-truth mesh. We also evaluated our method with only six input frames. As shown in Figure~\ref{Fig:synthetic}(g)(h), we are able to reconstruct the human model with quite sparse frames.

\begin{figure*}[ht]
	\centering
	\includegraphics[width=0.9999\linewidth]{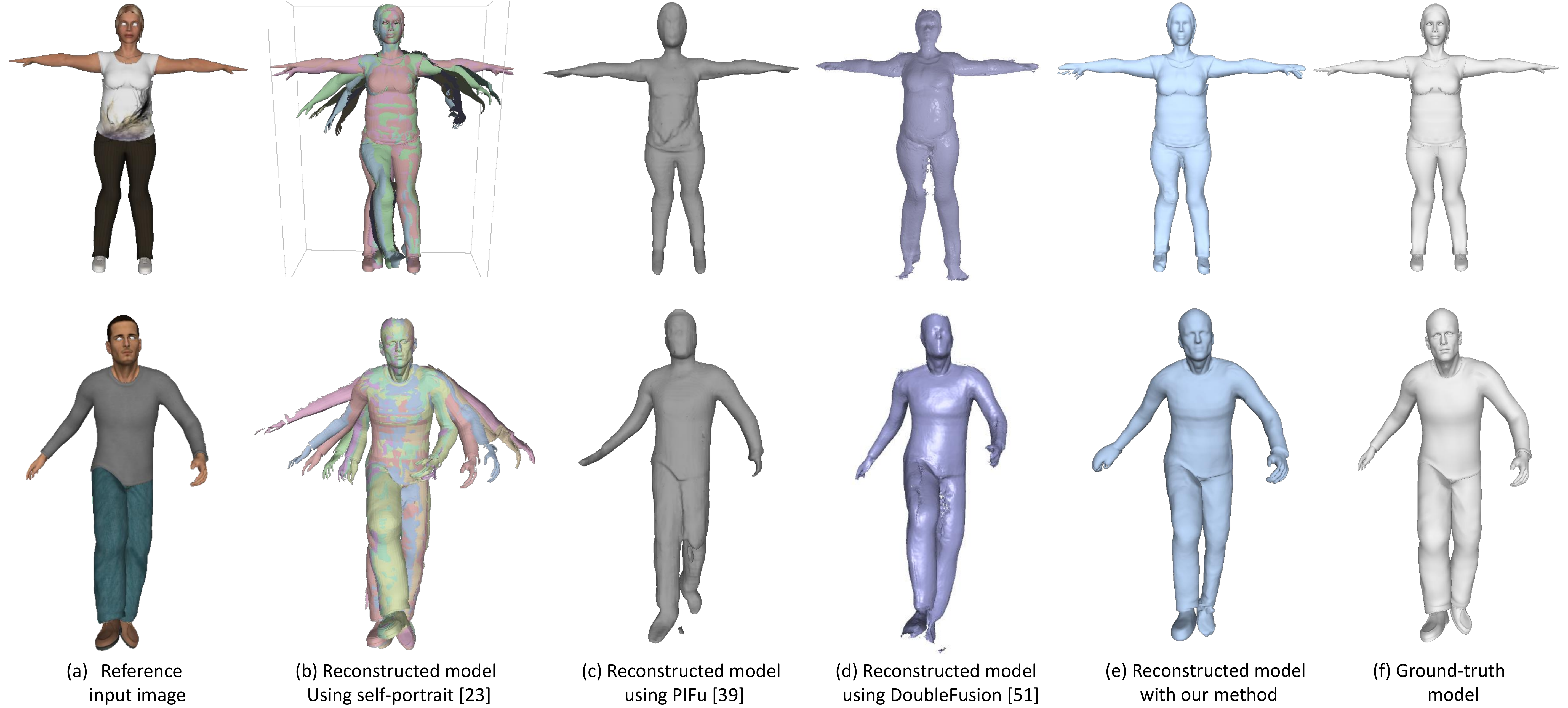}
	\caption{Comparison with state-of-the-art human body modeling methods on a synthetic dataset.}
	\label{Fig:synthetic_cmp}
\end{figure*}

3D self-portrait~\cite{Hao13selfportrait}, which also takes eight partial pieces as input, is closely related to our work. We implement the 3D self-portrait and test their method on our synthetic dataset. As can be seen in Figure~\ref{Fig:synthetic_cmp}(b), it is quite difficult to align those partial pieces without dealing with the large pose changes. Therefore, the misalignment appears especially around the arms and legs.

We also compare our method with the current state-of-the-art human body reconstruction method using deep learning techniques~\cite{saito2019pifu}. It is quite convenient to use a single color image as input, however, the reconstructed model is over-smoothed and lacks surface details as shown in Figure~\ref{Fig:synthetic_cmp}(c). Also the inherent depth ambiguity results in inaccurate 3D poses and body shapes.

Besides, to compare our method with the current state-of-the-art fusion based approach ~\cite{Yu18doublefusion} which fuses a depth sequence into a canonical model by continuously tracking the surface evolution, we have rendered a depth sequence with 90 frames for each human subject. To maintain continuous motion along the sequence, we conduct extrapolation among the selected sparse models. As shown in Figure~\ref{Fig:synthetic_cmp}(d), there are some artifacts along the legs and arms in the fused canonical models caused by the accumulated error and imperfect initialization as they require an A-pose as the starting pose. 

\begin{figure*}[!h]
	\centering
	\includegraphics[width=0.9999\linewidth]{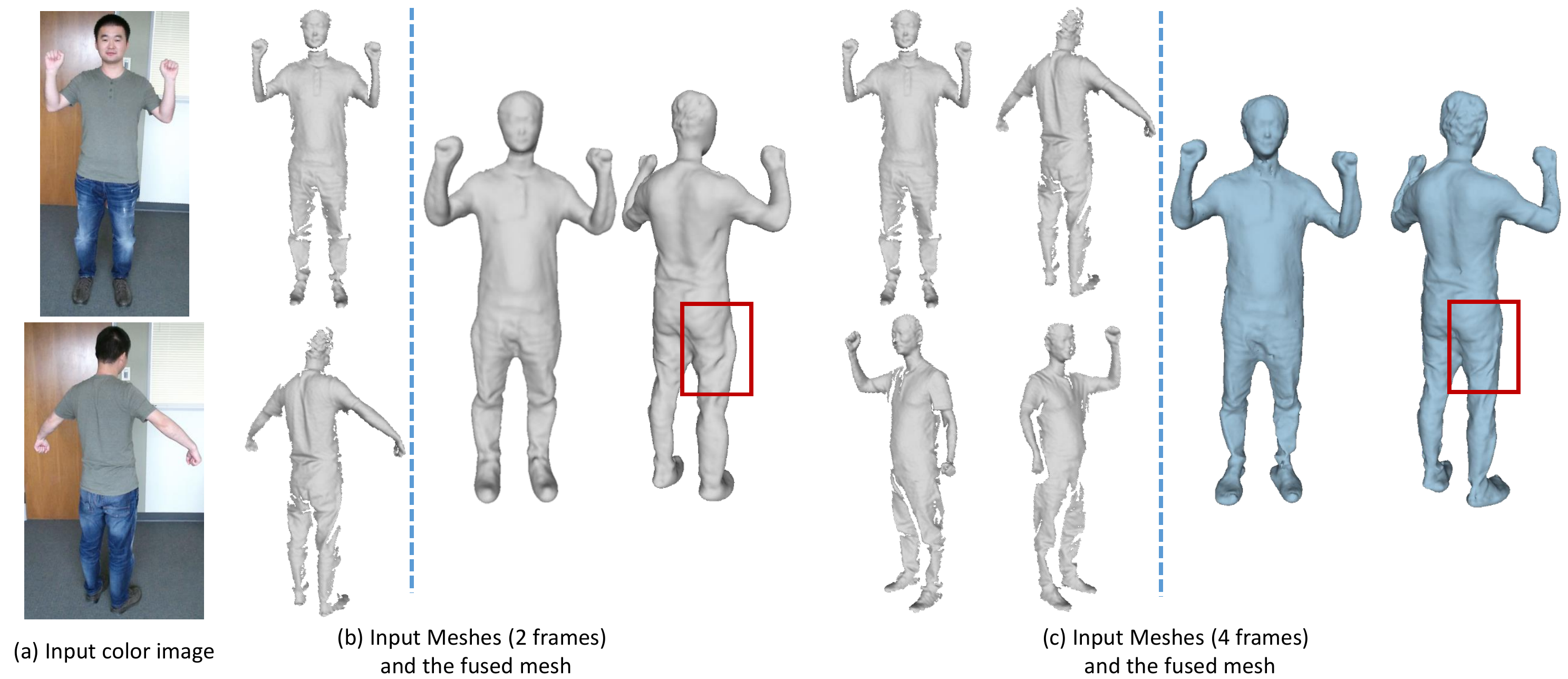}
	\caption{Demonstration of the reconstructed models with quite sparse frames. (a) shows the sampled color images. (b) shows the fused model and the two pieces used to reconstruct the model. The number of the vertices for the fused model is 44673. (c) shows the fused model and the four pieces used to reconstruct the model. The reconstructed model has 47540 vertices.}
	\label{Fig:framesNum}
\end{figure*}

Table~\ref{tab:error} shows the reconstruction error. We evaluate the reconstruction error of the fused models using our method with 1, 6 and 8 frames as input. For the reconstruction using only one frame, we take the optimized SMPL model as the reconstructed model. We also compute the reconstruction error for the models achieved from DoubleFusion~\cite{Yu18doublefusion} and PIFu~\cite{saito2019pifu}. As demonstrate in Table~\ref{tab:error}, our proposed method has achieved the best performance with reconstruction error as low as several millimeters. 

\begin{table}[!h]
\vspace{5pt}
\caption{Reconstruction error. For each human subject, we compute the distance of every vertex on the reconstructed model to its nearest vertex on the ground-truth model. The reconstruction error (in mm) indicates the average distance for all the vertices.}
\label{tab:error}
\begin{center}
\begin{tabular}{cccccc}
\hline
\multirow{2}{*}{\begin{tabular}[c]{@{}l@{}}human\\ subjects\end{tabular}} & PIFu~\cite{saito2019pifu} & DoubleFusion~\cite{Yu18doublefusion} &\multicolumn{3}{c}{ours (number of frames)} \\ \cline{4-6} 
                        & & & 1         & 6         & 8         \\ \hline 
subject 1                                                           &16.8  & 16.4   & 17.1     & 9.2    & \textbf{7.4}  \\ 
subject 2                                                           &68.1  & 18.9   & 19.2      & 10.3   & \textbf{8.7}   \\ 
subject 3                                                           &62.1  & 15.4  & 16.9     & 10.4   & \textbf{8.2}    \\ 
subject 4                                                           &58.1  & 51.7  & 18.7      & 9.6    & \textbf{6.8}   \\ 
mean error                                                          &51.28 & 25.63  & 17.98     & 9.88   & \textbf{7.78} \\ \hline
\end{tabular}
\end{center}
\end{table}

\subsection{Qualitative evaluation on real datasets}
For the qualitative evaluation, we have captured RGBD sequences of several human subjects with a Microsoft Kinect V2. The results of our method are displayed in Figure~\ref{Fig:real}. For each reconstruction, we use twelve RGBD frames as input. We take a frontal piece as the canonical frame and deform all other pieces onto it. As demonstrated in Figure~\ref{Fig:real}, complete human models with sufficient surface details are recovered. Besides, we can also deform the reconstructed human model onto any input scan. 

\begin{figure}[h]
	\centering
	\includegraphics[width=0.9999\linewidth]{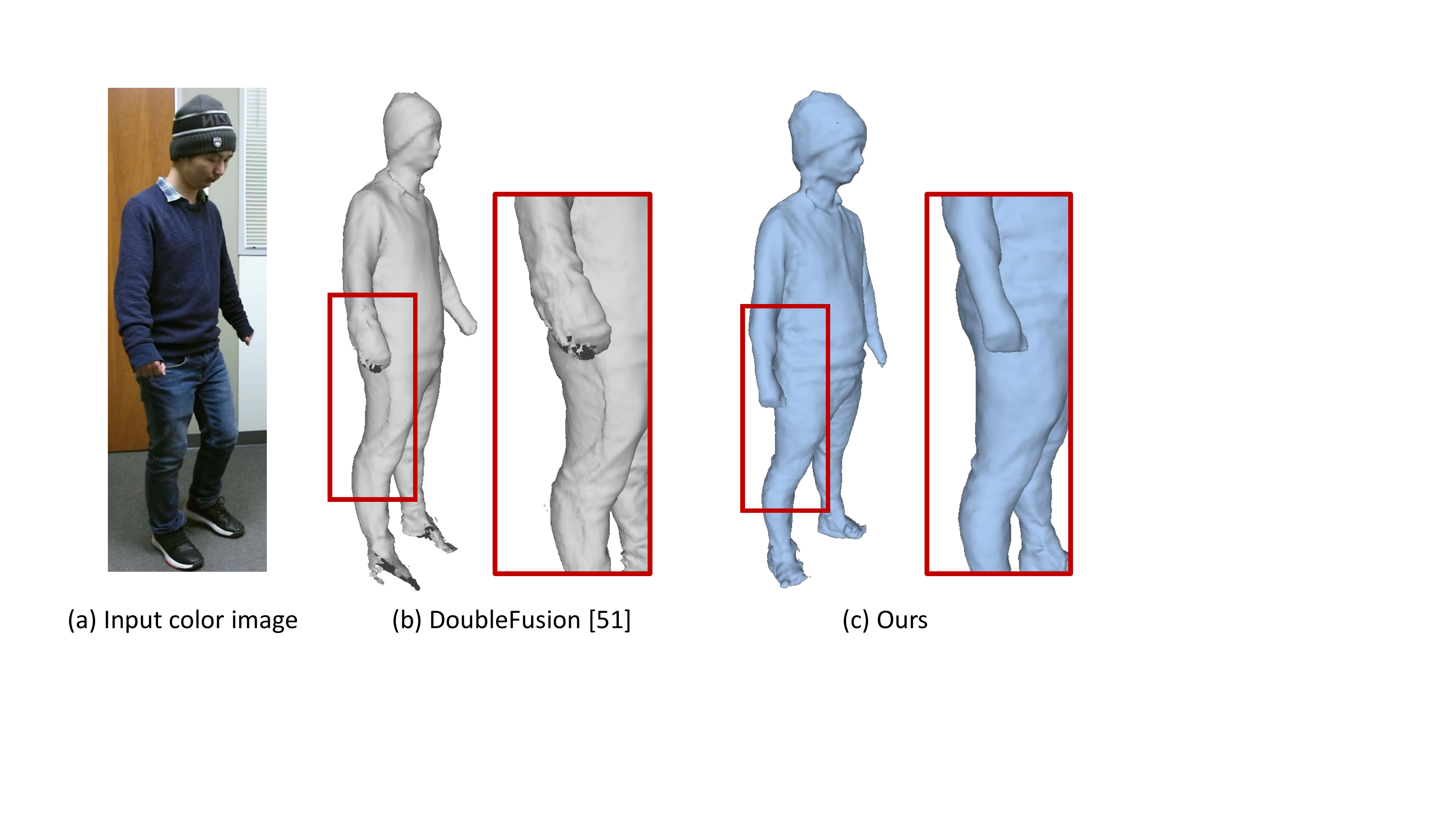}
	\caption{Comparison with dynamic fusion approach.}
	\label{Fig:doubleF}
\end{figure}

We have also conducted visual evaluation with DoubleFusion~\cite{Yu18doublefusion} on a real dataset with the results shown in Figure~\ref{Fig:doubleF}. The human subject was required to try to maintain A-pose while rotating in front of the camera. Although the DoubleFusion method also exploits human template to track the human poses, there are still some artifacts in the fused model caused by accumulated error as shown in Figure~\ref{Fig:doubleF}(b). The reason is that they rely on accurate tracking along the whole sequence. As compared with the fusion method, our proposed method is able to reconstruct complete models without any seams(Figure~\ref{Fig:doubleF}(c)).


In Figure~\ref{Fig:framesNum}, we demonstrate the ability of our method on model fusion with quite limited frames. In this case, since the overlapping regions between every two pieces are very small, it is not sufficient to perform pairwise registration. Therefore, we deform every partial scan onto the canonical space as guided by the SMPL template in the first place. We show the reconstructed models with only 2 and 4 pieces. The reconstruction becomes better when more frames are used. As shown in the red box, there are irregular bumps in the reconstructed model after Possion Surface Reconstruction when we take 2 pieces as input. The reconstructed surface gets better when we have 2 more pieces.

\begin{figure}[h]
	\centering
	\includegraphics[width=0.9999\linewidth]{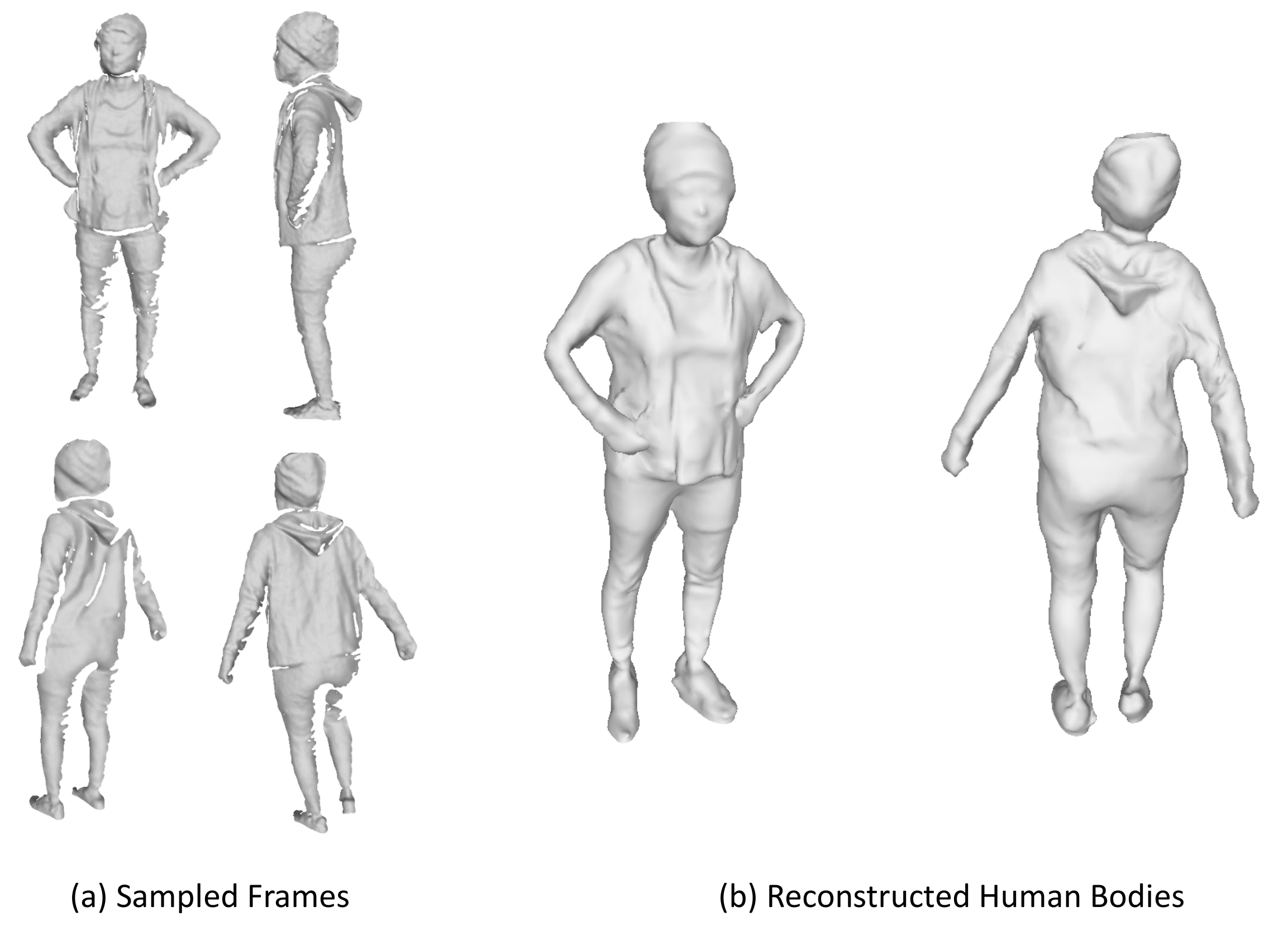}
	\caption{Results on changing topology. The reconstructed model has 58417 vertices.}
	\label{Fig:topology-rec}
\end{figure}

We further demonstrate the effectiveness of our method on dealing with topology changes in Figure~\ref{Fig:topology-rec}. We have tackled this problem explicitly while performing deformation, therefore we are able to generate pleasant results in this case.

\subsection{Applications on animation}
\begin{figure}[h]
	\centering
	\includegraphics[width=0.9999\linewidth]{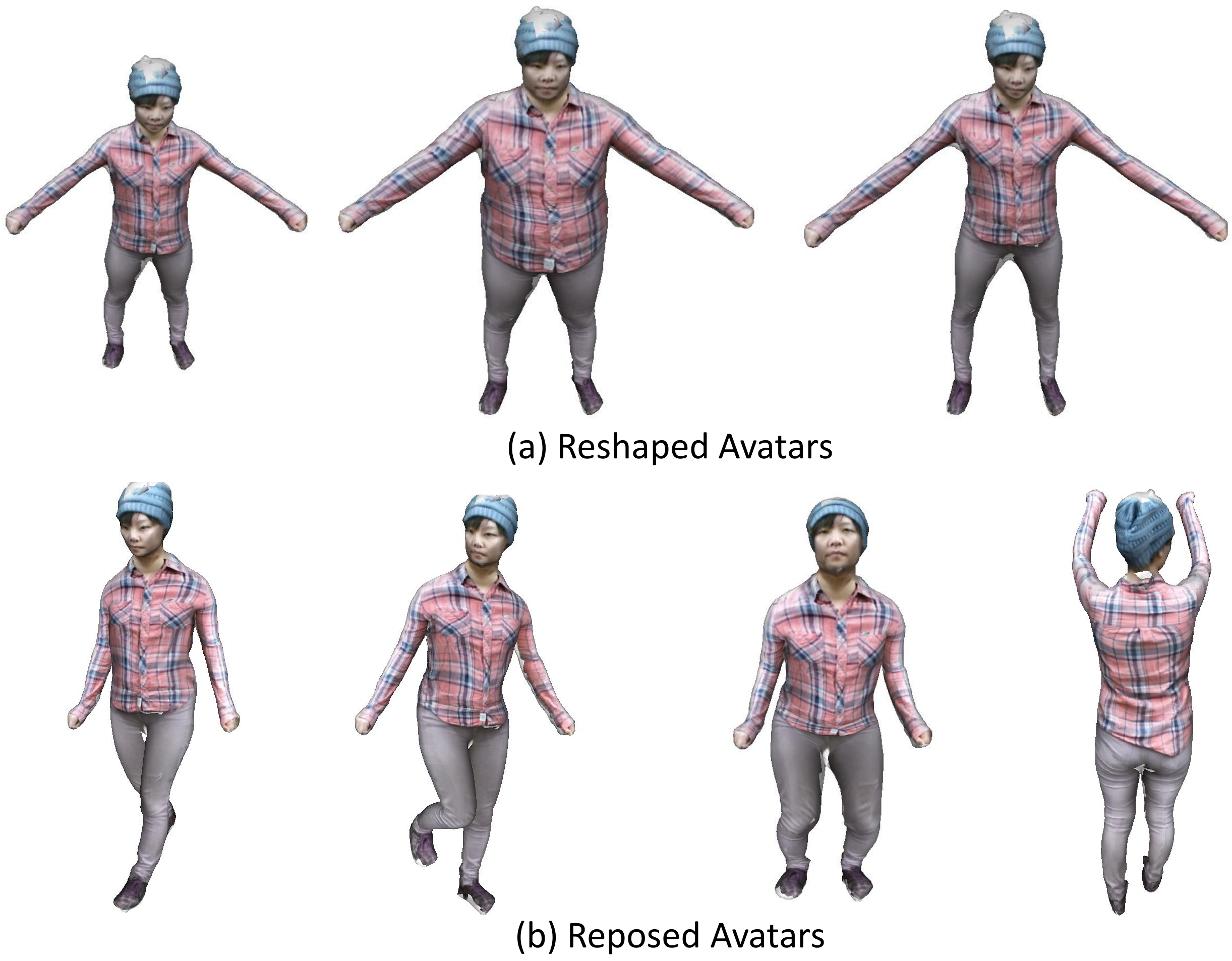}
	\caption{Reshaping and reposing of a human avatar.}
	\label{Fig:animate}
\end{figure}

In this section, we show some results on animated human avatars by building up personalized SMPL model. We could adjust the parameters representing the shape of the model to synthesize human models that are shorter/taller, or fatter/thinner as shown in Figure~\ref{Fig:animate}(a). Meanwhile, we could generate human avatars under various poses (as displayed in Figure~\ref{Fig:animate}(b)) by manipulating the pose parameters of our personalized SMPL model.

\subsection{Limitations}
\begin{figure}[h]
	\centering
	\includegraphics[width=0.9999\linewidth]{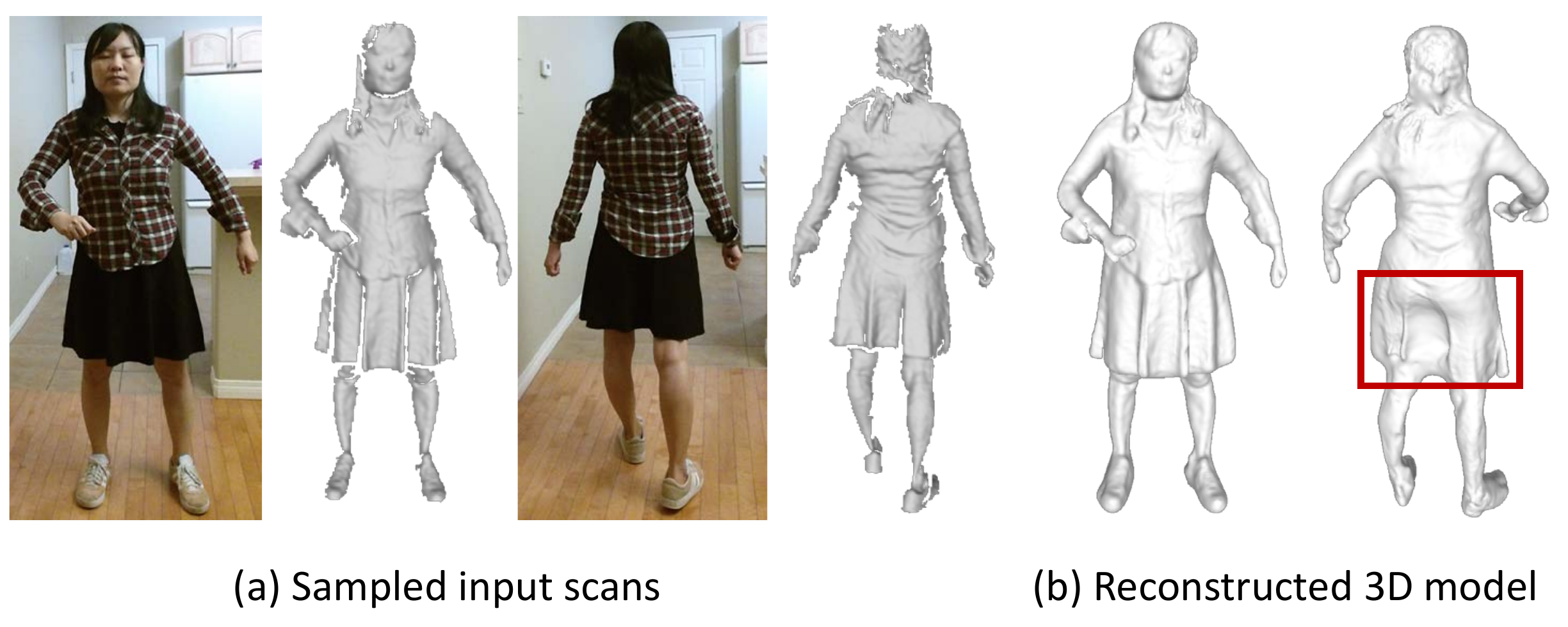}
	\caption{Reconstruction results on human subject with loose clothes. (a) shows two sampled input RGBD scans. (b) shows the reconstructed model from our approach. The red box highlights the artifacts on the reconstructed model where the shape of the dress was successfully recovered.}
	\label{Fig:failure}
\end{figure}

In this section, a failure case is demonstrated in Figure~\ref{Fig:failure} where the captured human subject is wearing a dress. During the pairwise registration we exploit the SMPL based human template to find initial correspondences between partial scans. Since this human template is built up from naked human models, it fails to find reliable matches around the folds of the dress. Eventually, we get the reconstructed model where the shape of the dress is not fully recovered. But we can still achieve reasonable results overall where the upper body and the legs are well reconstructed. It is noticed that there are also some artifacts around the hair, as it is quite noisy in the captured depth map for the hair due to its reflection characteristics.

\section{Conclusion and Future work}

In this paper, we have proposed a novel approach to build up a complete human avatar from only sparse RGBD images. To align those partial pieces of a human body under different poses and viewpoints into a canonical model, a SMPL based human template was utilized to align the input partial pieces. After constructing the complete human model, we presented a texture mapping method to construct spatially consistent texture maps for the reconstructed human model. Experiments on both synthetic and real datasets demonstrate the excellent performance (with reconstruction error in few millimeters) of our framework in reconstructing complete human bodies. As a potential application, animations are carried out with our reconstructed human avatar across various shapes and poses.

At the moment, the human modeling method is designed for a single person. For future work, we look at the more challenging problem of reconstructing multiple human subjects with interactions, which often contain significant occlusions and convoluted topological structures. 

\ifCLASSOPTIONcaptionsoff
\newpage
\fi


\bibliographystyle{ieee}
\bibliography{BIBREF}

%
%


\begin{IEEEbiography}
	[{\includegraphics[width=1in,height=1.25in,clip,keepaspectratio]{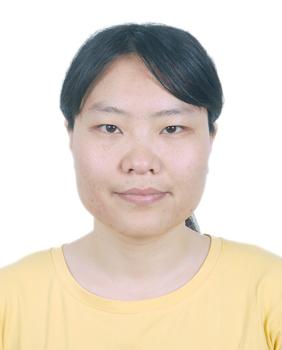}}]
	{Xinxin Zuo} received the M.E. degree from Northwestern Polytechnical University and Ph.D. degree from the University of Kentucky. She is currently a Postdoctoral Fellow at University of Alberta. Her research interests include computer vision and graphics, especially on 3D reconstruction and human modeling.
\end{IEEEbiography}

\begin{IEEEbiography}
	[{\includegraphics[width=1in,height=1.25in,clip,keepaspectratio]{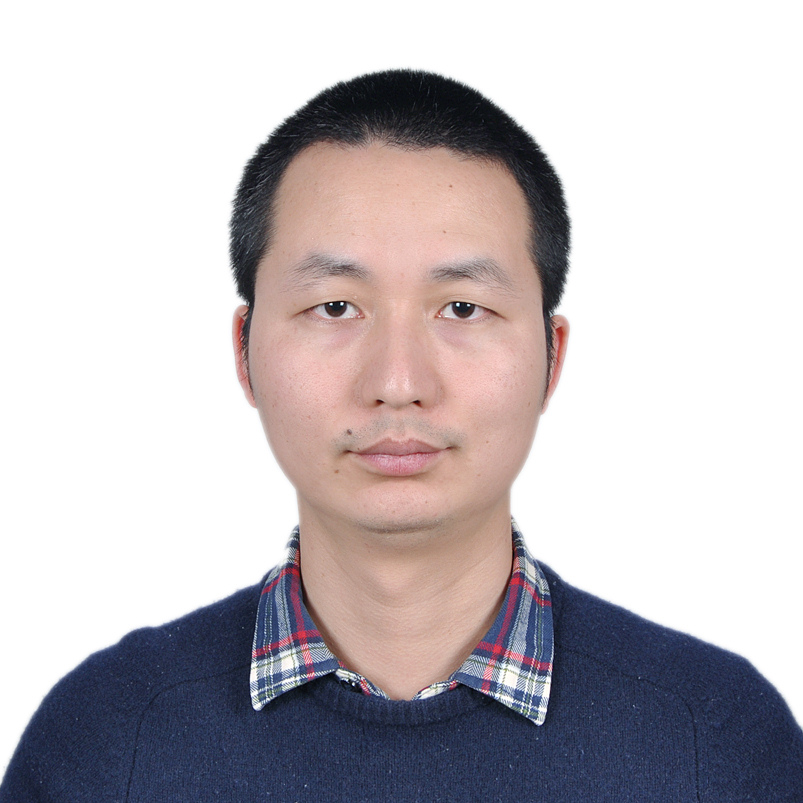}}]
	{Sen Wang} received the B.E. degree and Ph.D. degree from Northwestern Polytechnical University. From 2015 to 2016, he was a Visiting Ph.D. Student at the University of Kentucky. He is currently a Postdoctoral Fellow at University of Alberta. His research interests include computer vision and robotics. 
\end{IEEEbiography}

\begin{IEEEbiography}
	[{\includegraphics[width=1in,height=1.25in,clip,keepaspectratio]{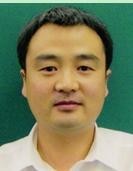}}]
	{Jiangbin Zheng} received the Ph.D. degree from Northwestern Polytecnical University, in 2002, where he is a Full Professor and Dean with school of Software. His research interests include computer graphics, computer vision and multimedia. He has published over 100 papers in the above related research area.
\end{IEEEbiography}

\begin{IEEEbiography}
	[{\includegraphics[width=1in,height=1.25in,clip,keepaspectratio]{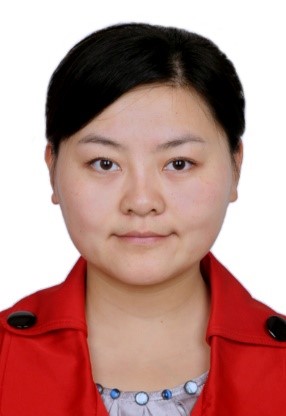}}]{Weiwei Yu}
	had been enrolled in "Sino-France Doctoral School" in 2006. She received the Ph.D. degree in navigation, guidance and control from Northwestern Polytechnical University in 2010. She received the PhD. degree in information science from Paris EST University in 2011. She is now an associate professor in the School of Mechatronics Engineering, Northwestern Polytechnical University. Her research expertise is mainly in artificial intelligence and bio-robot.
\end{IEEEbiography}

\begin{IEEEbiography}
	[{\includegraphics[width=1in,height=1.25in,clip,keepaspectratio]{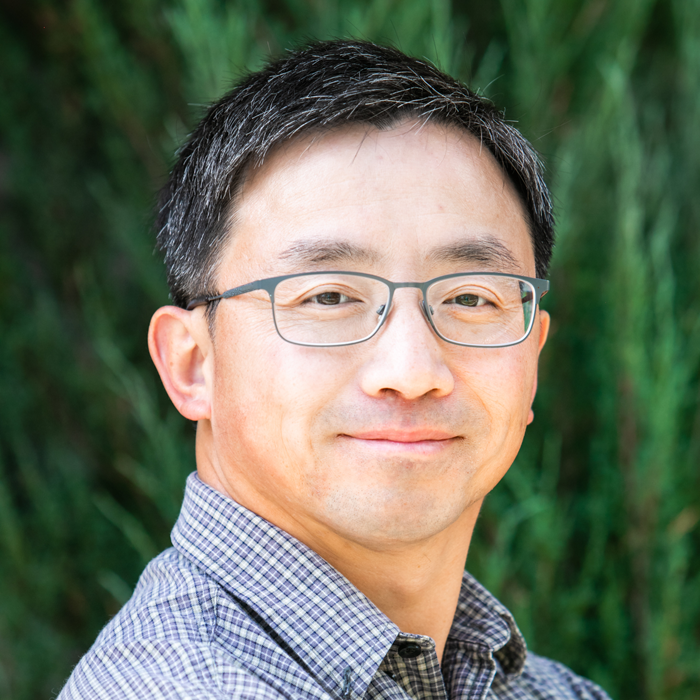}}]{Minglun Gong} is a Professor and Director at the School of Computer Science, University of Guelph. Before he joined Guelph in 2019, he was a Professor and Head at the Department of Computer Science, Memorial University of Newfoundland. He obtained his Ph.D. from the University of Alberta in 2003, his M.Sc. from the Tsinghua University in 1997, and his B.Engr. from the Harbin Engineering University in 1994. Dr. Gong’s research interests cover various topics in the broad area of visual computing (including computer graphics, computer vision, visualization, image processing, and pattern recognition). So far, he has published 130+ referred technical papers in journals and conference proceedings, including 20+ articles in ACM/IEEE transactions.
\end{IEEEbiography}

\begin{IEEEbiography}
	[{\includegraphics[width=1in,height=1.25in,clip,keepaspectratio]{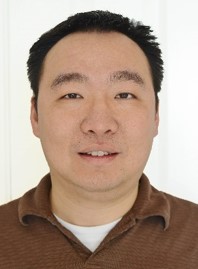}}]{Ruigang Yang (SM'13)}
	received the M.S. degree from Columbia University and the Ph.D. degree from the University of North Carolina at Chapel Hill. He is currently a Full Professor in computer science with the University of Kentucky. He has published over 100 papers, which, according to Google Scholar, has received over 10000 citations with an h-index of 50 (as of 2018). His research interests include computer graphics and computer vision, in particular in 3D reconstruction and 3D data analysis. He has received a number of awards, including the US NSF Career Award, in 2004 and the Dean’s Research Award from the University of Kentucky, in 2013. He is currently a senior member of IEEE.
\end{IEEEbiography}

\begin{IEEEbiography}
	[{\includegraphics[width=1in,height=1.25in,clip,keepaspectratio]{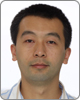}}]{Li Cheng} received the Ph.D. degree in computer science from the University of Alberta, Canada. He is an associate professor with the Department of Electrical and Computer Engineering, University of Alberta. Prior to coming back to University of Alberta, He has worked at A*STAR, Singapore, TTI-Chicago, USA, and NICTA, Australia. His research expertise is mainly on computer vision and machine learning. He is a senior member of the IEEE.
\end{IEEEbiography}

\ifCLASSOPTIONcaptionsoff
  \newpage
\fi

\end{document}